\documentclass[sigconf]{acmart}

\AtBeginDocument{%
  }

\copyrightyear{2026}
\acmYear{2026}
\setcopyright{cc}
\setcctype{by}

\acmConference[MM '26]
  {Proceedings of the 34th ACM International Conference on Multimedia}
  {November 10--14, 2026}
  {Rio de Janeiro, Brazil}

\acmBooktitle{Proceedings of the 34th ACM International Conference on
Multimedia (MM '26), November 10--14, 2026, Rio de Janeiro, Brazil}

\acmDOI{10.1145/3767308.3836545}
\acmISBN{979-8-4007-2213-4/2026/11}

\usepackage{colortbl}
\usepackage[ruled,linesnumbered]{algorithm2e}

\usepackage{multirow}

\begin{document}

\title{CODE: Cross-Modal Calibration and Dynamic Suppression for Open World Object Detection}

\author{Hao Xu}

\affiliation{%
  \institution{Beijing Institute of Technology}
  \city{Beijing}
  \country{China}
}
\email{xuhao\_cs@bit.edu.cn}

\author{Zhaoning Shi}
\affiliation{%
  \institution{Beijing Institute of Technology}
  \city{Beijing}
  \country{China}
}
\email{3220235133@bit.edu.cn}

\author{Hehe Jin}
\affiliation{%
  \institution{Beijing Institute of Technology}
  \city{Beijing}
  \country{China}
}
\email{1120223715@bit.edu.cn}

\author{Bo Ma}
\correspondingauthor
\affiliation{%
  \institution{Beijing Institute of Technology}
  \city{Beijing}
  \country{China}
}
\email{bma000@bit.edu.cn}

\renewcommand{\shortauthors}{Hao Xu et al.}


\begin{abstract}

Open World Object Detection (OWOD) built on multimodal foundation models often suffers from semantic ambiguity caused by unidirectional text-to-vision matching, while rigid outlier penalties may over-suppress unknown objects near known-class decision boundaries. We propose \textbf{CODE} (\textit{\underline{C}ross-Modal Calibrati\underline{O}n and \underline{D}ynamic Suppr\underline{E}ssion}), a unified inference-time framework with three complementary components. \textbf{Cross-Modal Joint Confidence Calibration} injects global visual prototypes to calibrate text-driven known-class predictions. \textbf{Uncertainty-Guided Universal Objectness Enhancement} measures classification hesitation from local visual responses to strengthen potential unknown objects. \textbf{Dynamic Outlier Suppression via Confidence Margin} replaces rigid suppression with a margin-aware adjustment that preserves ambiguous out-of-distribution instances. Experiments on the Real-World Detection benchmark demonstrate that,
with the OWL-ViT L/14 backbone, CODE achieves $21.7$ U-mAP and $40.8$
K-mAP in Task~1, surpassing the previous state of the art by $2.6$ and
$2.3$ points, respectively.
\end{abstract}

\ccsdesc[500]{Computing methodologies~Object detection}
\ccsdesc[300]{Computing methodologies~Neural networks}

\keywords{Open World Object Detection, Multi-modal Foundation Models, Cross-modal Inference, Out-of-Distribution Detection}

\maketitle

\section{Introduction}

Object detection, a cornerstone of computer vision, has traditionally operated under a ``closed-world'' assumption where the categories encountered during inference are strictly limited to those seen during training \cite{joseph2021towards, zohar2023open}. However, real-world environments are inherently dynamic and unpredictable, frequently presenting novel objects that fall outside pre-defined taxonomies \cite{li2024open, joseph2021towards}. Open World Object Detection (OWOD) has emerged to address this challenge, requiring models to accurately identify known categories while simultaneously localizing and flagging unknown objects without conflating them with the background or known classes \cite{joseph2021towards, zohar2023open, yang2025detecting}.

\begin{figure}[t]
  \centering

  \includegraphics[width=0.95\linewidth]{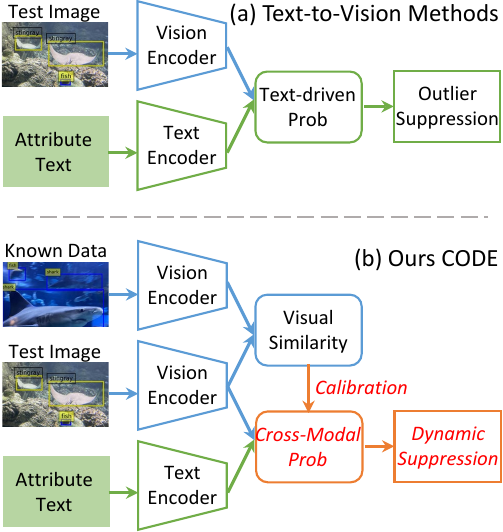} 
  
  \caption{Paradigm comparison between attribute-based OWOD methods and
CODE. Existing methods derive text-driven probabilities followed by
fixed outlier suppression, whereas CODE introduces known-class visual
references for cross-modal calibration and dynamic suppression.}
  
  \Description{A paradigm comparison between existing attribute-driven
open-world object detection methods and CODE. The existing method
independently encodes the test image and attribute text, computes a
text-driven probability, and applies fixed outlier suppression. CODE
additionally encodes known-category visual data, computes visual
similarities for cross-modal probability calibration, and applies
dynamic suppression before producing the final prediction.}
  
  \label{fig:intro}
\end{figure}

The advent of multimodal foundation models, such as
CLIP~\cite{radford2021learning} and
OWL-ViT~\cite{minderer2022simple}, has introduced rich open-vocabulary
knowledge learned from large-scale image--text pairs into OWOD.
To bridge the domain gap between generic pre-training and downstream
applications, recent methods employ fine-grained, class-agnostic
attributes as intermediate semantic representations
\cite{zohar2023open,yang2025detecting,xi2024umb}.

As illustrated in Figure~\ref{fig:intro}(a), existing attribute-driven
methods independently encode the test image and attribute descriptions,
derive text-driven probabilities, and subsequently apply an outlier
suppression mechanism. This unidirectional text-to-vision process lacks
a direct visual reference when textual attributes are insufficient or
visually ambiguous. In contrast, CODE introduces visual embeddings
obtained from known-category data to calibrate the text-driven
probabilities and replaces rigid suppression with a dynamic mechanism,
as shown in Figure~\ref{fig:intro}(b).

Despite their potential, existing attribute-based OWOD paradigms face three limitations. First, unidirectional text-to-vision matching can produce \emph{semantic ambiguity}: discriminative visual attributes may become unnoticeable under occlusion, blur, viewpoint changes, or poor illumination, while textual attributes alone may be insufficiently specific to distinguish hard objects from visually similar known categories or background regions. Second, existing methods often estimate universal objectness from the maximum attribute response, which cannot reliably separate ambiguous unknown objects from background distractors. Finally, the rigid Maximum Concept Matching (MCM) mechanism~\cite{ming2022delving} may over-penalize boundary unknown objects. Because these objects can share attributes with several known categories, a suppression term determined only by the maximum known-class probability may incorrectly remove them as background.

To overcome these challenges, we propose \textbf{CODE} (Cross-Modal Calibration and Dynamic Suppression), a unified framework for cross-modal joint inference. Our method introduces three core components designed to balance the precision of known targets with the robust recall of unknowns. First, we implement \textbf{Cross-Modal Joint Confidence Calibration (CMJCC)}, which explicitly injects global visual prototypes to provide a robust visual-to-visual reference, effectively calibrating text-driven confidences. Second, we design the \textbf{Uncertainty-Guided Universal Objectness Enhancement (UGUOE)} module. By quantifying classification ``hesitation'' via the variance of local visual responses, UGUOE actively boosts the response of potential unknown objects located in ``semantic vacuums''. 
Finally, we introduce \textbf{Dynamic Outlier Suppression via Confidence Margin (DOSCM)}, which replaces rigid penalties with a margin-based adjustment. By evaluating the exclusivity of known predictions through the probability gap between the primary and secondary Softmax responses, DOSCM protects ambiguous out-of-distribution objects from over-suppression.

We evaluate our framework on the Real-World Detection (RWD)
benchmark~\cite{zohar2023open}, which spans five diverse scenarios:
Aquatic, Aerial, Game, Medical, and Surgery. CODE achieves the best
overall performance under both OWL-ViT backbones. Notably, with the
L/14 backbone, CODE improves Task~1 U-mAP and K-mAP by $2.6$ and $2.3$
points over the previous state of the art, respectively, demonstrating
that unknown-object discovery can be improved without sacrificing
known-class recognition.

The main contributions of this work are summarized as follows:
\begin{itemize}
\item We propose \textbf{Cross-Modal Joint Confidence Calibration (CMJCC)}, which injects class-level visual prototypes as visual-to-visual references to calibrate text-driven known-class predictions and alleviate semantic ambiguity.
\item We introduce \textbf{Uncertainty-Guided Universal Objectness Enhancement (UGUOE)}, which estimates local classification hesitation from the distribution of visual prototype responses and enhances potential unknown objects that receive weak or ambiguous textual activation. 
\item We develop \textbf{Dynamic Outlier Suppression via Confidence Margin (DOSCM)}, which uses the probability gap between the two most likely known classes to dynamically regulate outlier suppression. Together, the three components form a lightweight inference-time framework that improves both known- and unknown-object detection across diverse real-world scenarios.
\end{itemize}

\section{Related Works}

\subsection{Open-World Object Detection with Multi-modal Foundation Models}

Open World Object Detection (OWOD) aims to alleviate the strict closed-set assumptions of standard detectors, requiring models to simultaneously identify known classes and discover unseen objects \cite{joseph2021towards}. Early pioneering works, such as ORE \cite{joseph2021towards}, OW-DETR \cite{gupta2022ow}, PROB \cite{zohar2023prob}, and others \cite{li2024open, xi2024ktcn, zhao2023revisiting, liu2024uoa, liu2025adaptive, liu2025detr}, primarily rely on pseudo-labeling or contrastive clustering to define decision boundaries. Recently, the rise of multi-modal foundation models has catalyzed Open-Vocabulary Object Detection (OVOD), which leverages rich visual-language knowledge for zero-shot generalization. Representative OVOD methods include ViLD \cite{gu2021open}, RegionCLIP \cite{zhong2022regionclip}, OWL-ViT \cite{minderer2022simple}, and F-VLM \cite{kuo2022f}, among others \cite{zareian2021open, yao2023detclipv2, maaz2022class}. Although OVD models demonstrate remarkable zero-shot generalization to novel concepts, they rely entirely on explicit text prompts (i.e., category names) and fundamentally fail to handle the core OWOD challenge: detecting "unknowns" without any prior category information.

To bridge this gap, an attribute-driven OWOD paradigm has recently emerged as the state-of-the-art. Methods such as FOMO \cite{zohar2023open}, UMB \cite{xi2024umb}, and PASS \cite{yang2025detecting} utilize Large Language Models to decompose abstract categories into fine-grained, class-agnostic attributes (e.g., shape, texture, material). By learning these intermediate semantic representations, these models identify unknown objects that share visual or functional characteristics with known base classes. Although these methods enhance the discriminability of novel targets and provide explainability, they still struggle to detect objects characterized by semantic ambiguity in attribute mapping. Consequently, we propose \textbf{CMJCC} to resolve mapping ambiguities and \textbf{UGUOE} to facilitate robust unknown discovery.

\textbf{Open-ended object detection.}
Recent open-ended detection methods extend open-vocabulary detection beyond a fixed query set through region-language pre-training or bidirectional visual-semantic alignment. GenerateU~\cite{lin2024generative} employs generative region-language modeling to discover open-ended categories, whereas Open-Det~\cite{cao2025open} improves visual-language alignment through a dedicated training framework. In contrast, CODE targets attribute-driven OWOD and performs lightweight inference-time calibration using pre-computed textual attributes and visual prototypes, without online language-model decoding or additional bidirectional detector training.

\subsection{Cross-modal OOD Detection and Unknown Probability Modeling}

A fundamental challenge in OWOD is the evaluation of an object's Out-of-Distribution (OOD) score, or "unknownness," to effectively distinguish it from known categories and background noise. Traditional unimodal OOD detection often relies on post-hoc scores calculated from the internal activations of a trained vision model. Representative approaches include Maximum Softmax Probability (MSP) \cite{hendrycks2016baseline}, energy-based scores \cite{liu2020energy}, and MaxLogit \cite{hendrycks2019scaling}, as well as other established techniques such as ODIN \cite{liang2017enhancing}, ReAct \cite{sun2021react}, and Grad-Norm \cite{huang2021importance}, among others \cite{fort2021exploring, koner2021oodformer}. While these methods provide a statistical basis for identifying outliers, they are primarily confined to a single modality and often struggle with the complex semantic shifts inherent in open-world environments.

With the integration of multi-modal representations, cross-modal OOD evaluation has emerged as a powerful alternative. Maximum Concept Matching (MCM) \cite{ming2022delving} establishes a zero-shot OOD baseline by aligning visual features with known textual concepts through scaled softmax. Subsequent works have further expanded this boundary: CLIPN \cite{wang2023clipn} introduces negative prompts to explicitly model the rejection of known classes, while PEFT-MCM \cite{ming2024does} utilizes parameter-efficient fine-tuning to refine the OOD decision boundary, supported by additional cross-modal studies \cite{esmaeilpour2022zero, maaz2022class, zhou2022conditional, xu2024tgtrack}. Furthermore, recent Dual-Pattern Matching (DPM) \cite{zhang2024vision} frameworks fuse textual similarity with visual distribution patterns to bridge the modality gap. Despite their efficacy, these methods often employ rigid suppression strategies that lead to the over-suppression of boundary unknown targets. To mitigate this, we propose \textbf{DOSCM}, which adaptively protects ambiguous out-of-distribution instances by assessing the exclusivity of predictions through probability margins.

\section{Methodology}

\begin{figure*}[t]
  \centering
  \includegraphics[width=\textwidth]{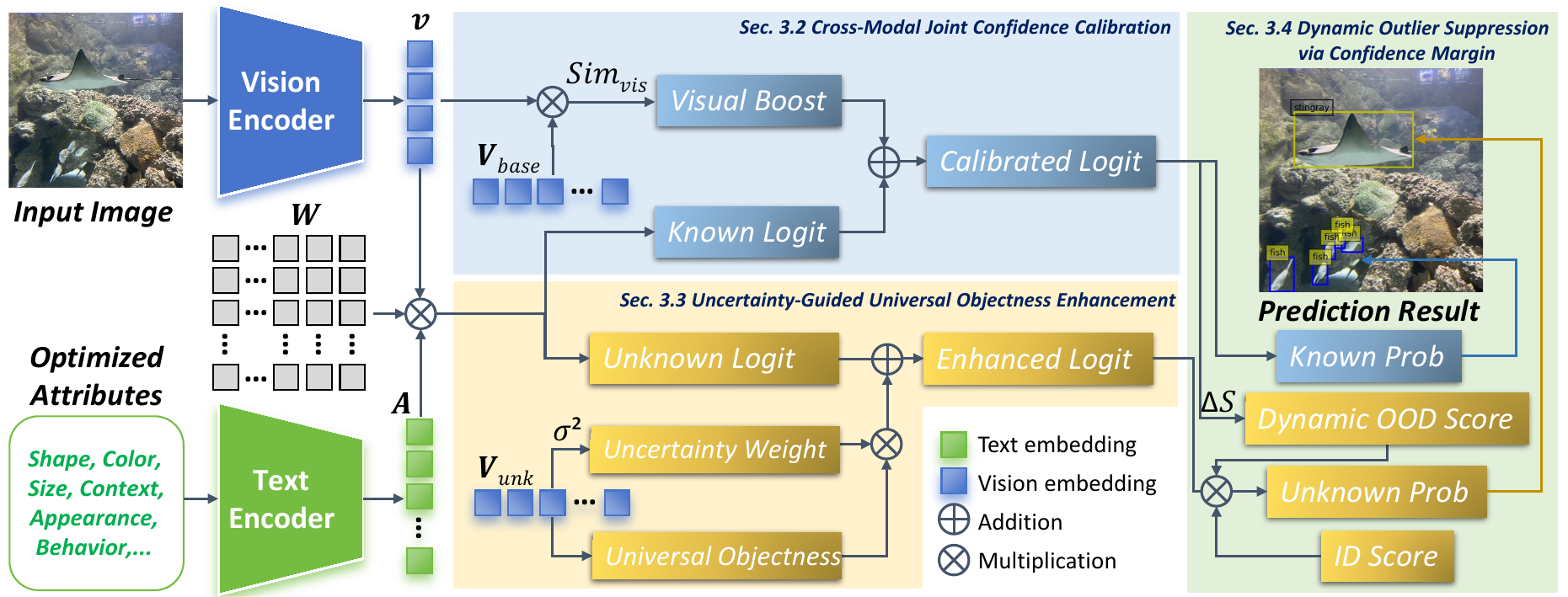} 
  
  \caption{Inference architecture of CODE. CMJCC calibrates known logits
with cached visual prototypes, UGUOE enhances unknown logits from visual
uncertainty, and DOSCM converts confidence margins into dynamic OOD
scores.}
  
  \Description{The inference architecture of CODE. An input image is
encoded into candidate visual embeddings, while optimized textual
attributes are encoded and mapped to initial known and unknown logits.
The CMJCC branch compares candidate embeddings with pre-computed
known-class visual prototypes to calibrate known logits. The UGUOE
branch combines visual-response uncertainty with universal objectness
to enhance unknown logits. The DOSCM branch uses the confidence margin
of calibrated known predictions to compute a dynamic out-of-distribution
score and produce the final known and unknown probabilities.}
  
  \label{fig:overall_framework}
\end{figure*}

Our proposed framework, \textbf{CODE} (Cross-Modal Calibration and
Dynamic Suppression), is built upon the attribute-based OWL-ViT
baseline~\cite{minderer2022simple}. As illustrated in
Figure~\ref{fig:overall_framework}, CODE uses pre-computed optimized
attribute embeddings and known-class visual prototypes. CMJCC
calibrates text-derived known logits, UGUOE enhances the unknown logit
from local visual-response dispersion, and DOSCM adjusts outlier
suppression using the top-1/top-2 confidence margin. Attribute
generation and prototype construction are both completed offline and
are not invoked during test-time inference.

\subsection{Preliminary: Attribute-based OWOD Baseline}

Following the prevalent OWOD paradigms based on multi-modal foundation models (e.g., OWL-ViT~\cite{minderer2022simple}), our framework utilizes the attribute semantic space as an intermediate representation to map visual features of candidate regions to category labels.

\textbf{Attribute Selection and Adaptation.} Given $K$ known categories, the model first prompts a Large Language Model (LLM) to generate a rich set of fine-grained attribute descriptions. These generated attributes are processed by a text encoder to form an initial attribute pool. To extract and adapt a highly discriminative subset from this redundant pool, previous methods ~\cite{yang2025detecting} optimize the attribute embeddings $\boldsymbol{A}$ and the mapping weights $\boldsymbol{W}$. During the training phase, this is achieved by minimizing a comprehensive loss function comprising four components:
\begin{equation}
\mathcal{L} = \mathcal{L}_{CE} + \lambda_{POT} \mathcal{L}_{POT} + \lambda_{MSE} \mathcal{L}_{MSE} + \lambda_{L1} \mathcal{L}_{L1}
\end{equation}
Here, $\mathcal{L}_{CE}$ is the standard cross-entropy classification loss. $\mathcal{L}_{POT}$ represents the partial optimal transport distance used to select the most relevant attributes. $\mathcal{L}_{MSE}$ is the Mean Squared Error loss applied to align the text-derived attributes with the average vision-derived class embeddings, which ensures stable refinement under limited supervision. Finally, $\mathcal{L}_{L1}$ applies L1 regularization to the mapping weights $\boldsymbol{W}$ to encourage sparsity.

\textbf{Logits Representation.}
After selection and adaptation, we obtain $N$ optimized attribute
embeddings,
$\boldsymbol{A}=\{\boldsymbol{a}_n\}_{n=1}^{N}$,
where $N$ denotes the fixed number of attributes retained after
attribute selection and adaptation. Its value is specified in the
implementation details. We define the classification mapping matrix as
$\boldsymbol{W}\in\mathbb{R}^{N\times(K+1)}$, where the first $K$
columns correspond to the known categories and the last column
corresponds to the semantic projection of the unknown category.
For the visual embedding $\boldsymbol{v}$ of a candidate region,
its unified logits $Logit\in\mathbb{R}^{K+1}$ are computed as:

\begin{equation}
Logit = [Logit_{K, 1}, \dots, Logit_{K, K}, Logit_{U}] = \text{sim}(\boldsymbol{v}, \boldsymbol{A}) \cdot \boldsymbol{W}
\end{equation}
where $\text{sim}(\cdot)$ denotes the cosine similarity. Thus, the text-driven logits for known categories are denoted as $Logit_{K}$, and the corresponding component for the unknown category is $Logit_{U}$.

\textbf{Inference.} During the inference stage, the baseline method computes the classification probabilities for known and unknown targets separately:

1. \textbf{Known Target Probability ($P_K$)}: By applying the Sigmoid activation function $\sigma(\cdot)$ to the known components, the confidence score of the candidate region belonging to a known class $k$ is obtained:
\begin{equation}
P_K(\boldsymbol{v}) = \sigma(Logit_{K, k})
\end{equation}

2. \textbf{Unknown Target Probability ($P_U$)}: The determination of an unknown target relies not only on its classification logit $Logit_{U}$ but is also modulated by task relevance ($p_{ID}$) and unknownness ($p_{OOD}$):
\begin{equation}
P_U(\boldsymbol{v}) = \sigma(Logit_{U}) \cdot p_{ID} \cdot p_{OOD}
\end{equation}
where $p_{ID}$ measures the matching degree between the visual features and any known attributes:
\begin{equation}
p_{ID} = \max_{n \in \{1, \dots, N\}} \sigma(\boldsymbol{a}_n^{\top} \boldsymbol{v})
\end{equation}
and $p_{OOD}$ implements out-of-distribution outlier suppression based on the Maximum Concept Matching (MCM) mechanism \cite{ming2022delving}. Instead of using the independent classification probabilities, MCM penalizes the unknown score using the highest normalized probability derived from applying the Softmax function to the known logits:
\begin{equation}
p_{OOD} = 1 - \max_{k \in \{1, \dots, K\}} \text{Softmax}(Logit_{K})_k
\end{equation}

\subsection{Cross-Modal Joint Confidence Calibration}

As analyzed in Sec. 1, the text-driven logits $Logit_{K}$ derived from unidirectional attribute mapping are susceptible to semantic ambiguity, particularly when visual attributes are partially occluded or the visual-textual alignment is weak. To address this limitation, we propose the \textbf{Cross-Modal Joint Confidence Calibration (CMJCC)} framework. Unlike previous methods that rely solely on textual descriptions, CMJCC explicitly incorporates global visual prototypes to provide a robust visual-to-visual reference, effectively calibrating the confidence of known categories.

\textbf{Visual Prototype Calculation.}
For each dataset, we construct a visual prototype bank offline using
the frozen OWL-ViT visual encoder. For a known category $k$, let
$\mathcal{D}_k=\{\boldsymbol{v}_{k,i}\}_{i=1}^{M_k}$ denote the set of
$M_k$ region embeddings extracted from the ground-truth boxes of class
$k$ in the labeled training or support split. The visual prototype
$\boldsymbol{p}_k$ is defined as the normalized centroid of these
embeddings:
\begin{equation}
\boldsymbol{p}_k =
\operatorname{Norm}\left(
\frac{1}{M_k}
\sum_{i=1}^{M_k}\boldsymbol{v}_{k,i}
\right)
\end{equation}

The prototypes are computed and cached before inference and are never
updated using test images. In Task~1, prototypes are constructed
strictly from annotated known categories, without using unknown-category
labels. In Task~2, prototypes of previously known categories remain
frozen, while prototypes of newly introduced categories are computed
from their available labeled samples and appended to the prototype bank.

The collective visual base matrix is then constructed as $\boldsymbol{V}_{base} = [\boldsymbol{p}_1, \dots, \boldsymbol{p}_K]^{\top} \in \mathbb{R}^{K \times D}$. This prototype injection ensures that the model captures the intrinsic visual distribution of each class, compensating for potential gaps in textual attribute coverage.

\textbf{Visual Similarity and Gating.}
For a candidate region $\boldsymbol{v}_i$, we compute its direct
visual-level similarities to the known-category prototypes as
\begin{equation}
Sim_{vis,i}
=
\boldsymbol{v}_i\boldsymbol{V}_{base}^{\top}
\in\mathbb{R}^{K}.
\end{equation}

To suppress background and irrelevant visual responses, we compute a
candidate-specific activation threshold:
\begin{equation}
\tau_i
=
\max\left(
\frac{1}{K}
\sum_{k=1}^{K}
[Sim_{vis,i}]_k
+
m,\,
\tau_{min}
\right),
\label{eq:dynamic_threshold}
\end{equation}
where $[Sim_{vis,i}]_k$ is the similarity between candidate
$\boldsymbol{v}_i$ and prototype $k$, $m$ is a fixed scalar margin,
and $\tau_{min}$ is the minimum activation floor. Since the mean is
computed only over the $K$ prototypes of the current candidate,
$\tau_i$ is independent of the composition and size of the inference
batch.

The corresponding visual boost is obtained by applying a ReLU gate:
\begin{equation}
Logit_{boost,i}
=
\operatorname{ReLU}
\left(
Sim_{vis,i}-\tau_i
\right).
\label{eq:visual_boost}
\end{equation}
For notational simplicity, we omit the candidate index $i$ in the
following equations.

\textbf{Joint Logit Calibration.} To ensure that the visual boost is consistent with the logit distribution of the pre-trained foundation model, we introduce an absolute scaling factor $S_{scale}$, also referred to as the reverse temperature. Following the intrinsic temperature setting $\tau_{OWL-ViT}$ of the foundation model, the scaling factor is defined as:
\begin{equation}
S_{scale} = \frac{1}{\tau_{OWL-ViT}}
\end{equation}
The final calibrated logits for known categories, $Logit_{K}^{calib}$, are formulated by fusing the original text-driven logits with the visually-guided boost component:
\begin{equation}
Logit_{K}^{calib} = Logit_{K} + \alpha \cdot S_{scale} \cdot Logit_{boost}
\end{equation}
where $\alpha$ is a modulation coefficient that balances the contribution of the visual prior. By mapping the visual similarity back to a sharp decision probability gradient through $S_{scale}$, CMJCC preserves the discriminative power of the original feature space while providing the necessary cross-modal calibration.

\subsection{Uncertainty-Guided Universal Objectness Enhancement}

While the calibration mechanism in Sec. 3.2 improves the reliability of known classes, the detection of unknown objects remains challenging due to the lack of explicit supervision. Previous methods~\cite{yang2025detecting, zohar2023open, xi2024umb} often rely on a generic objectness score derived from the maximum attribute response, which fails to distinguish between background noise and ambiguous unknown targets. To bridge this gap, we propose an \textbf{Uncertainty-Guided Universal Objectness Enhancement (UGUOE)} module. By analyzing the local distribution of visual similarities, UGUOE quantifies the model's classification uncertainty to generate a robust universal objectness prior, actively boosting the response of potential unknown objects.

\textbf{Local Visual Response Aggregation.} For a candidate region $\boldsymbol{v}$, we consider its Top-$K_{u}$ nearest visual prototypes from $\boldsymbol{V}_{base}$ to capture its local semantic neighborhood. Let $S_{local} = \{s_1, s_2, \dots, s_{K_{u}}\}$ be the set of the $K_{u}$ largest similarity values in $Sim_{vis}$. The base visual-driven unknown response $\mu_{unk}$ is defined as the average of these local similarities:
\begin{equation}
\mu_{unk} = \frac{1}{K_{u}} \sum_{s \in S_{local}} s
\end{equation}
A high $\mu_{unk}$ indicates that the region exhibits strong object-like visual characteristics, even if it does not strictly align with a single known category.

\textbf{Uncertainty Modeling via Variance.} A key observation is that for a clearly defined known object, its visual similarity will be highly concentrated on a specific prototype, resulting in a high variance within $S_{local}$. Conversely, an unknown object often exhibits ambiguous, low-variance responses across multiple related known prototypes. We quantify this "hesitation" or classification uncertainty using the unbiased variance $\sigma^{2}(S_{local})$:
\begin{equation}
\sigma^{2}(S_{local}) = \frac{1}{K_{u}-1} \sum_{s \in S_{local}} (s - \mu_{unk})^2
\end{equation}
We then formulate the uncertainty-guided weight $W_{unc}$ using an exponential decay function to penalize high-certainty (known) regions and favor low-variance (ambiguous) candidates:
\begin{equation}
W_{unc} = \max\Big(w_{min}, \min\big(w_{max}, \exp(-\gamma \cdot \sigma^{2}(S_{local}))\big)\Big)
\end{equation}
where $\gamma$ controls the sensitivity to uncertainty, and $[w_{min}, w_{max}]$ clips the weight to a stable range.

\textbf{Enhanced Universal Objectness and Logit Fusion.} To extract the pure universal objectness activation, we first apply a thresholding function to the base visual response:
\begin{equation}
Obj_{unk} = \text{ReLU}(\mu_{unk} - \tau_{unk})
\end{equation}
where $\tau_{unk}$ is the activation threshold used to filter out background distractors. The final enhanced logit for the unknown category, $Logit_{U}^{enh}$, is then obtained by injecting this uncertainty-weighted universal objectness into the original text-driven unknown logit $Logit_{U}$:
\begin{equation}
Logit_{U}^{enh} = Logit_{U} + \beta \cdot S_{scale} \cdot W_{unc} \cdot Obj_{unk}
\end{equation}
where $\beta$ is a scaling coefficient. This formulation ensures that objects located in the "semantic vacuum" between known classes receive a significant universal objectness boost, thereby increasing their probability $P_U(\boldsymbol{v})$ before the outlier suppression phase.

\subsection{Dynamic Outlier Suppression via Confidence Margin}

Through the CMJCC and UGUOE modules, we have obtained the calibrated known probabilities $P_K^{calib}(\boldsymbol{v})$ and the enhanced unknown logits $Logit_{U}^{enh}$. The final step in the OWOD inference pipeline is to determine the Out-of-Distribution (OOD) outlier score $p_{OOD}$ to distinguish genuine unknown objects from known categories. 

\textbf{Revisiting the MCM Bottleneck.} As formulated in Sec. 3.1, the standard MCM mechanism utilizes the highest Softmax-normalized probability for suppression: $p_{OOD} = 1 - \max(\text{Softmax}(Logit_{K}))$. However, this rigid penalty strategy introduces a severe flaw at the decision boundary. A high Top-1 probability in the Softmax distribution does not necessarily guarantee absolute exclusivity; if the Top-2 probability is also high, the model is actually experiencing semantic confusion between two known classes. In such boundary scenarios, the region is highly likely to be a hard unknown instance sharing attributes with multiple known categories. The standard MCM blindly applies a massive penalty, erroneously suppressing these hard unknowns into the background.

\textbf{Confidence Margin Calculation.} To address this over suppression, we propose the \textbf{Dynamic Outlier Suppression via Confidence Margin (DOSCM)}. We argue that strong suppression should only be triggered when the model exhibits absolute, exclusive certainty about a single known class. To quantify this exclusivity, we extract the Top-1 and Top-2 probabilities from the Softmax-normalized distribution of the calibrated logits, denoted as $\mathcal{S} = \text{Softmax}(Logit_{K}^{calib})$:
\begin{equation}
S_{top1} = \max(\mathcal{S}), \quad S_{top2} = \max(\mathcal{S} \setminus \{S_{top1}\})
\end{equation}
We then define the confidence margin $\Delta S$ as the gap between the most confident and the second most confident predictions:
\begin{equation}
\Delta S = S_{top1} - S_{top2}
\end{equation}
A large $\Delta S$ indicates exclusive certainty towards a specific known class, whereas a small $\Delta S$ signifies boundary ambiguity.

\textbf{Dynamic Suppression and Final Inference.} We modulate the baseline MCM penalty using this confidence margin to compute the effective suppression score $MCM_{eff}$:
\begin{equation}
MCM_{eff} = S_{top1} \cdot \Delta S
\end{equation}
Consequently, the dynamic OOD suppression score $p_{OOD}^{dyn}$ is formulated as:
\begin{equation}
p_{OOD}^{dyn} = 1 - MCM_{eff}
\end{equation}
Finally, substituting $p_{OOD}^{dyn}$ and the enhanced unknown logit $Logit_{U}^{enh}$ back into the baseline inference framework, the ultimate probability for the unknown target is derived as:
\begin{equation}
P_U^{final}(\boldsymbol{v}) = \sigma(Logit_{U}^{enh}) \cdot p_{ID} \cdot p_{OOD}^{dyn}
\end{equation}
By decoupling the absolute maximum Softmax response from the penalty mechanism and introducing the margin-based dynamic adjustment, DOSCM effectively protects ambiguous unknown objects located in the semantic margins, thereby significantly improving the recall of hard out-of-distribution targets without compromising the precision of known classes.

\section{Experiments}

\subsection{Experimental Setup}

\subsubsection{Datasets}

We evaluate CODE on the Real-World Detection (RWD)
benchmark~\cite{zohar2023open}, which contains five practical
scenarios: \textbf{Aquatic}, \textbf{Aerial}, \textbf{Game},
\textbf{Medical}, and \textbf{Surgery}. Following the official RWD
protocol, each dataset is evaluated in two stages. Task~1 measures
unknown-object discovery from annotated known categories, whereas
Task~2 evaluates Previously Known (PK) and Currently Known (CK)
categories after the model is jointly retrained using all samples
available at that stage. Thus, Task~2 represents an expanded-category
evaluation rather than a strictly sequential incremental-learning
setting. We follow the official category splits, image partitions, and
evaluation protocol.

\subsubsection{Evaluation Metrics}

To comprehensively measure the balance between the accurate identification of known classes and the robust recall of unknown classes, we employ the following quantitative metrics:
\begin{itemize}
    \item \textbf{K-mAP and U-mAP:} In Task 1, we calculate the mean Average Precision (mAP) for Known and Unknown categories, respectively.
    \item \textbf{PK-mAP and CK-mAP:} In Task 2, we evaluate the detection performance on Previously Known (PK) categories and Currently Known (CK) categories newly introduced in the task.
\end{itemize}
Unlike early OWOD works that primarily focused on Recall for unknown objects, we strictly adopt mAP as the primary metric. This ensures a more rigorous evaluation of the detector's classification accuracy and its ability to suppress background noise.

\subsubsection{Implementation Details}

\textbf{Architecture and Environment.}
All experiments are conducted on NVIDIA GeForce RTX 4090 GPUs using
PyTorch. We employ frozen OWL-ViT B/16 and L/14 models as the multimodal
foundation backbones. Following FOMO~\cite{zohar2023open}, GPT-3.5 is
used only offline to generate the initial fine-grained attribute
descriptions. The resulting attributes are encoded and cached before
inference, and GPT-3.5 is not invoked during test-time detection.
The known-category visual prototypes are also pre-computed and cached
before inference according to the protocol described in Sec.~3.2.

\textbf{Hyperparameter Settings.}
The number of retained attribute embeddings is set to $N=25$ for all
datasets. In CMJCC, the scalar margin for visual similarity gating is
set to $m=0.2$, and the minimum activation floor is
$\tau_{min}=0.4$. Following the original OWL-ViT
configuration~\cite{minderer2022simple}, the temperature coefficient is
fixed at $\tau_{OWL\text{-}ViT}=0.07$, corresponding to
$S_{scale}\approx14.28$, while the cross-modal calibration coefficient
is set to $\alpha=0.5$.

In UGUOE, we retain the top $K_u=3$ local visual responses. The
uncertainty sensitivity factor is set to $\gamma=100.0$, and the
universal objectness threshold is set to $\tau_{unk}=0.25$. For
numerical stability, the uncertainty weight $W_{unc}$ is clamped to
$[w_{min},w_{max}]=[0.1,1.0]$. During attribute optimization, the
linear projection layers and calibration weights are optimized using
AdamW. Unless otherwise stated, the same hyperparameter configuration
is used across the five RWD datasets.

\begin{table*}[t]
  \caption{OWOD results on the five RWD datasets. We report U- and K-mAP
in Task~1 and PK- and CK-mAP in Task~2. Results of previous methods are
taken from their original papers, while CODE uses our final unified
configuration. The $\Delta$ rows report CODE minus PASS; positive
values indicate improvements. $^\dagger$ GT baselines use ground-truth
class names to detect unknown objects and serve as an open-vocabulary
upper reference.}
  \label{tab:main_results}
  \centering
  \resizebox{\textwidth}{!}{
  \begin{tabular}{l|cc|cc||cc|cc||cc|cc||cc|cc||cc|cc||cc|cc}
    \toprule
    \textbf{Dataset ($\rightarrow$)} & \multicolumn{4}{c||}{\textbf{Aquatic}} & \multicolumn{4}{c||}{\textbf{Aerial}} & \multicolumn{4}{c||}{\textbf{Game}} & \multicolumn{4}{c||}{\textbf{Medical}} & \multicolumn{4}{c||}{\textbf{Surgery}} & \multicolumn{4}{c}{\textbf{Overall}} \\
    \midrule 
    \textbf{Task ID ($\rightarrow$)} & \multicolumn{2}{c|}{\textbf{Task 1}} & \multicolumn{2}{c||}{\textbf{Task 2}} & \multicolumn{2}{c|}{\textbf{Task 1}} & \multicolumn{2}{c||}{\textbf{Task 2}} & \multicolumn{2}{c|}{\textbf{Task 1}} & \multicolumn{2}{c||}{\textbf{Task 2}} & \multicolumn{2}{c|}{\textbf{Task 1}} & \multicolumn{2}{c||}{\textbf{Task 2}} & \multicolumn{2}{c|}{\textbf{Task 1}} & \multicolumn{2}{c||}{\textbf{Task 2}} & \multicolumn{2}{c|}{\textbf{Task 1}} & \multicolumn{2}{c}{\textbf{Task 2}} \\
    & \cellcolor{blue!10}U & \cellcolor{orange!10}K & \cellcolor{orange!10}PK & \cellcolor{orange!10}CK & \cellcolor{blue!10}U & \cellcolor{orange!10}K & \cellcolor{orange!10}PK & \cellcolor{orange!10}CK & \cellcolor{blue!10}U & \cellcolor{orange!10}K & \cellcolor{orange!10}PK & \cellcolor{orange!10}CK & \cellcolor{blue!10}U & \cellcolor{orange!10}K & \cellcolor{orange!10}PK & \cellcolor{orange!10}CK & \cellcolor{blue!10}U & \cellcolor{orange!10}K & \cellcolor{orange!10}PK & \cellcolor{orange!10}CK & \cellcolor{blue!10}U & \cellcolor{orange!10}K & \cellcolor{orange!10}PK & \cellcolor{orange!10}CK \\
    \midrule
    \multicolumn{25}{l}{\textit{B/16 Backbone:}} \\ 
    \rowcolor{gray!10} BASE-ZS+GT$^\dagger$ & 29.8 & 45.0 & 45.0 & 36.7 & 1.3 & 5.7 & 5.7 & 1.4 & 15.0 & 0.4 & 0.4 & 0.1 & 0.5 & 0.0 & 0.0 & 0.1 & 5.6 & 1.5 & 1.4 & 0.3 & 10.4 & 10.5 & 10.5 & 7.7 \\
    BASE-ZS & 6.2 & 45.0 & 45.0 & 36.7 & 0.9 & 5.7 & 5.7 & 1.4 & 15.7 & 0.4 & 0.4 & 0.1 & 0.2 & 0.0 & 0.0 & 0.1 & 1.4 & 1.5 & 1.4 & 0.3 & 4.9 & 10.5 & 10.5 & 7.7 \\
    BASE-ZS+IN & 26.5 & 45.1 & 45.1 & 36.7 & 1.9 & 5.7 & 5.7 & 1.4 & 2.4 & 0.3 & 0.3 & 0.0 & 0.6 & 0.0 & 0.0 & 0.1 & 1.7 & 1.4 & 1.0 & 0.3 & 6.6 & 10.5 & 10.4 & 7.7 \\
    BASE-ZS+LLM & 24.7 & 45.1 & 45.1 & 36.5 & 1.4 & 5.7 & 5.7 & 1.4 & 15.1 & 0.4 & 0.4 & 0.1 & 0.6 & 0.0 & 0.0 & 0.1 & 8.9 & 1.5 & 1.3 & 0.3 & 10.2 & 10.5 & 10.5 & 7.7 \\
    BASE-FS & 7.1 & 41.1 & 41.1 & 31.9 & 1.2 & 10.4 & 10.1 & 4.0 & 16.0 & 4.6 & 4.8 & 3.9 & 0.6 & 6.1 & 6.1 & 3.3 & 1.3 & 11.9 & 11.3 & 10.9 & 5.2 & 14.8 & 14.7 & 10.8 \\
    \midrule
    FOMO~\cite{zohar2023open} & 3.5 & 43.8 & 44.1 & 40.8 & 0.9 & 12.0 & 12.6 & 5.4 & 13.3 & 3.8 & 4.4 & 4.1 & 2.1 & 6.4 & 5.5 & 11.5 & 6.1 & 12.7 & 12.9 & 11.0 & 5.2 & 15.7 & 15.9 & 14.6 \\

    PASS~\cite{yang2025detecting}
& 5.2 & 43.4 & 43.2 & 46.6
& 1.9 & 14.0 & 16.0 & 7.0
& 21.5 & 10.0 & 7.7 & 9.0
& 4.9 & 8.4 & 6.8 & 12.1
& 14.3 & 15.6 & 13.1 & 14.7
& 9.6 & 18.3 & 17.4 & 17.9 \\

    CODE (Ours)
& 10.4 & 40.2 & 43.5 & 52.4
& 2.9 & 17.6 & 18.7 & 7.1
& 25.2 & 13.4 & 9.1 & 11.9
& 3.0 & 8.7 & 7.4 & 7.9
& 14.6 & 14.4 & 18.1 & 22.4
& \textbf{11.2} & \textbf{18.9}
& \textbf{19.4} & \textbf{20.3} \\
\rowcolor{green!5}
\textbf{\textcolor{green!60!black}{$\Delta$}}
& \textbf{\textcolor{green!60!black}{+5.2}}
& \textbf{\textcolor{green!60!black}{-3.2}}
& \textbf{\textcolor{green!60!black}{+0.3}}
& \textbf{\textcolor{green!60!black}{+5.8}}
& \textbf{\textcolor{green!60!black}{+1.0}}
& \textbf{\textcolor{green!60!black}{+3.6}}
& \textbf{\textcolor{green!60!black}{+2.7}}
& \textbf{\textcolor{green!60!black}{+0.1}}
& \textbf{\textcolor{green!60!black}{+3.7}}
& \textbf{\textcolor{green!60!black}{+3.4}}
& \textbf{\textcolor{green!60!black}{+1.4}}
& \textbf{\textcolor{green!60!black}{+2.9}}
& \textbf{\textcolor{green!60!black}{-1.9}}
& \textbf{\textcolor{green!60!black}{+0.3}}
& \textbf{\textcolor{green!60!black}{+0.6}}
& \textbf{\textcolor{green!60!black}{-4.2}}
& \textbf{\textcolor{green!60!black}{+0.3}}
& \textbf{\textcolor{green!60!black}{-1.2}}
& \textbf{\textcolor{green!60!black}{+5.0}}
& \textbf{\textcolor{green!60!black}{+7.7}}
& \textbf{\textcolor{green!60!black}{+1.6}}
& \textbf{\textcolor{green!60!black}{+0.6}}
& \textbf{\textcolor{green!60!black}{+2.0}}
& \textbf{\textcolor{green!60!black}{+2.4}} \\

    \midrule
    \midrule
    \multicolumn{25}{l}{\textit{L/14 Backbone:}} \\
    \rowcolor{gray!10} BASE-ZS+GT$^\dagger$ & 34.8 & 36.0 & 36.0 & 42.3 & 1.0 & 7.9 & 7.2 & 0.8 & 12.4 & 0.9 & 0.8 & 0.3 & 2.4 & 0.2 & 0.2 & 0.3 & 2.4 & 0.2 & 2.6 & 1.3 & 10.6 & 9.0 & 9.4 & 9.0 \\
    BASE-ZS & 0.7 & 35.9 & 36.0 & 42.3 & 9.1 & 8.2 & 7.2 & 0.8 & 6.8 & 0.9 & 0.8 & 0.3 & 0.0 & 0.2 & 0.2 & 0.3 & 3.6 & 2.9 & 2.6 & 1.3 & 4.1 & 9.6 & 9.4 & 9.0 \\
    BASE-ZS+IN & 19.6 & 35.8 & 35.8 & 41.8 & 2.3 & 7.2 & 6.9 & 0.9 & 15.8 & 0.9 & 0.8 & 0.3 & 0.9 & 0.1 & 0.1 & 0.2 & 3.1 & 2.1 & 1.9 & 1.1 & 8.3 & 9.2 & 9.1 & 8.8 \\
    BASE-ZS+LLM & 24.7 & 35.8 & 35.8 & 42.2 & 0.6 & 7.6 & 7.2 & 0.8 & 12.5 & 0.9 & 0.8 & 0.2 & 1.6 & 0.1 & 0.1 & 0.2 & 12.6 & 2.6 & 2.5 & 1.3 & 10.4 & 9.4 & 9.3 & 9.0 \\
    BASE-FS & 2.4 & 43.6 & 42.9 & 42.8 & 9.7 & 23.7 & 21.9 & 13.0 & 8.2 & 10.4 & 10.2 & 13.4 & 1.1 & 23.2 & 21.7 & 24.2 & 3.6 & 26.0 & 25.0 & 7.4 & 5.0 & 25.4 & 24.3 & 20.2 \\
    \midrule
    FOMO~\cite{zohar2023open} & 18.2 & 50.1 & 48.1 & 47.1 & 6.0 & 25.3 & 23.7 & 16.0 & 30.4 & 10.7 & 9.9 & 11.2 & 9.4 & 21.8 & 19.9 & 34.6 & 12.0 & 29.0 & 28.9 & 8.5 & 15.2 & 27.4 & 26.1 & 23.5 \\

    PASS~\cite{yang2025detecting}
& 21.7 & 53.9 & 56.6 & 58.3
& 8.4 & 34.2 & 36.1 & 20.2
& 36.0 & 24.3 & 23.7 & 26.3
& 13.1 & 34.3 & 30.0 & 32.0
& 16.6 & 45.6 & 47.9 & 43.3
& 19.1 & 38.5 & 38.9 & 36.0 \\

    CODE (Ours)
& 24.7 & 62.7 & 65.3 & 59.6
& 11.3 & 39.9 & 37.2 & 20.4
& 33.2 & 22.6 & 28.7 & 27.0
& 13.7 & 35.1 & 40.3 & 38.3
& 25.7 & 43.9 & 46.3 & 35.6
& \textbf{21.7} & \textbf{40.8}
& \textbf{43.6} & \textbf{36.2} \\
\rowcolor{green!5}
\textbf{\textcolor{green!60!black}{$\Delta$}}
& \textbf{\textcolor{green!60!black}{+3.1}}
& \textbf{\textcolor{green!60!black}{+8.8}}
& \textbf{\textcolor{green!60!black}{+8.7}}
& \textbf{\textcolor{green!60!black}{+1.3}}
& \textbf{\textcolor{green!60!black}{+2.9}}
& \textbf{\textcolor{green!60!black}{+5.7}}
& \textbf{\textcolor{green!60!black}{+1.1}}
& \textbf{\textcolor{green!60!black}{+0.2}}
& \textbf{\textcolor{green!60!black}{-2.8}}
& \textbf{\textcolor{green!60!black}{-1.7}}
& \textbf{\textcolor{green!60!black}{+5.0}}
& \textbf{\textcolor{green!60!black}{+0.7}}
& \textbf{\textcolor{green!60!black}{+0.6}}
& \textbf{\textcolor{green!60!black}{+0.8}}
& \textbf{\textcolor{green!60!black}{+10.3}}
& \textbf{\textcolor{green!60!black}{+6.3}}
& \textbf{\textcolor{green!60!black}{+9.1}}
& \textbf{\textcolor{green!60!black}{-1.7}}
& \textbf{\textcolor{green!60!black}{-1.6}}
& \textbf{\textcolor{green!60!black}{-7.7}}
& \textbf{\textcolor{green!60!black}{+2.6}}
& \textbf{\textcolor{green!60!black}{+2.3}}
& \textbf{\textcolor{green!60!black}{+4.7}}
& \textbf{\textcolor{green!60!black}{+0.2}} \\

    \bottomrule
  \end{tabular}
  }
\end{table*}

\textbf{Overall Performance.}
As shown in Table~\ref{tab:main_results}, CODE achieves the best
overall performance under both backbones. With OWL-ViT L/14, CODE
obtains $21.7$ U-mAP and $40.8$ K-mAP in Task~1, together with
$43.6$ PK-mAP and $36.2$ CK-mAP in Task~2. Compared with PASS, these
results correspond to improvements of \textbf{+2.6},
\textbf{+2.3}, \textbf{+4.7}, and \textbf{+0.2} points,
respectively. With B/16, CODE also improves the four overall metrics
by \textbf{+1.6}, \textbf{+0.6}, \textbf{+2.0}, and
\textbf{+2.4} points. The consistent improvements across the two
backbones demonstrate that CODE benefits both known-category
recognition and unknown-object discovery.

\textbf{Analysis of Task~1.}
CODE improves U-mAP on four of the five RWD scenarios. The largest
gain occurs on Surgery, where U-mAP increases by \textbf{+9.1}
points. It also improves Aquatic, Aerial, and Medical by
\textbf{+3.1}, \textbf{+2.9}, and \textbf{+0.6} points,
respectively. These gains support the effectiveness of UGUOE, which
uses the dispersion of local visual responses to activate potential
unknown objects that receive ambiguous known-category responses.

CODE also raises the overall K-mAP from $38.5$ to $40.8$, including
gains of \textbf{+8.8}, \textbf{+5.7}, and \textbf{+0.8} points
on Aquatic, Aerial, and Medical. This result indicates that CMJCC
compensates for semantic ambiguity by introducing direct visual
references from known categories. DOSCM further preserves boundary
unknowns by adjusting suppression according to the confidence margin,
thereby alleviating the conventional trade-off between known precision
and unknown discovery.

On Game, CODE decreases U-mAP and K-mAP by $2.8$ and $1.7$ points.
Our feature-space analysis gives a high average inter-class cosine
similarity of $0.82$ on this domain, supporting the interpretation that
some boundary unknowns are confused with nearby known-category
clusters.

\textbf{Analysis of Task~2.}
Since Task~2 jointly retrains the model using all samples available at
that stage, the results reflect its capacity to represent the expanded
category set rather than a strictly sequential incremental-learning
process. Under L/14, CODE improves the overall PK-mAP and CK-mAP by
\textbf{+4.7} and \textbf{+0.2} points. In particular, PK-mAP
increases by \textbf{+8.7} on Aquatic and \textbf{+10.3} on
Medical, with corresponding CK-mAP gains of \textbf{+1.3} and
\textbf{+6.3} points. CODE also improves both metrics on Aerial and
Game.

These gains demonstrate the importance of CMJCC in Task~2. When the
category space is expanded, the visual prototypes provide direct
class-level references that calibrate the text-driven representations
of both previously and currently known categories. On Surgery, CODE
remains below PASS in Task~2. Its high inter-class cosine similarity
of $0.87$ indicates that highly overlapping fine-grained categories
remain challenging even after visual calibration.

\subsection{Generalization to Standard Benchmarks}

We further evaluate CODE beyond RWD. On M-OWODB
Task~1~\cite{joseph2021towards}, CODE achieves $86.0$ U-Recall and
$70.0$ mAP, outperforming MAVL~\cite{maaz2022class}
($50.1$/$64.0$), DEUS~\cite{heo2026detecting}
($65.1$/$66.2$), and OrthogonalDet~\cite{sun2024exploring}
($24.6$/$61.3$), where each pair denotes U-Recall/mAP. On
LVIS~\cite{gupta2019lvis}, CODE obtains $32.0$
$\mathrm{AP}_{\mathrm{rare}}$, compared with $22.3$ for
GenerateU~\cite{lin2024generative} and $31.2$ for
Open-Det~\cite{cao2025open}. These results support generalization
beyond the small-category RWD scenarios. Because multimodal foundation
models may have encountered common benchmark categories during
pre-training, we retain RWD as the primary benchmark.

\begin{figure*}[t]
  \centering
  \includegraphics[width=\linewidth]{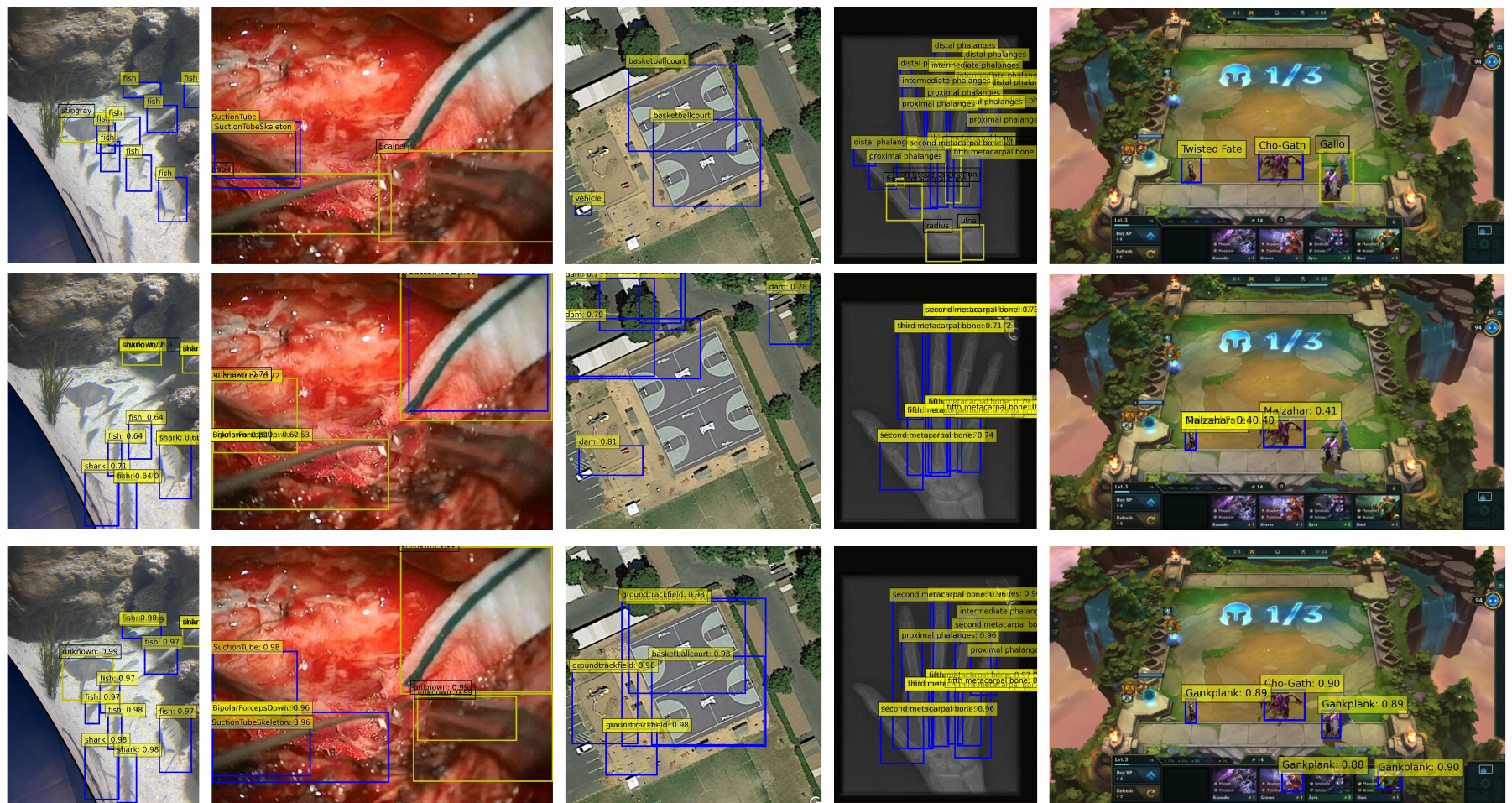}
  \caption{Qualitative comparison on five RWD datasets. Rows show ground
truth, PASS, and CODE; blue and yellow boxes denote known and unknown
objects, respectively. Columns show Aquatic and Surgery (Task~1), and
Aerial, Medical, and Game (Task~2).}
  \Description{A grid of 15 images arranged in 3 rows and 5 columns comparing object detection results. The columns represent five datasets: Aquatic, Surgery, Aerial, Medical, and Game. The top row shows the original images with ground truth bounding boxes. The middle row displays predictions from the PASS baseline, which visibly misses some objects or misclassifies them. The bottom row displays results from the proposed CODE framework, demonstrating bounding boxes that align much more accurately with the ground truth. Blue boxes indicate known objects and yellow boxes indicate unknown objects. Overall, the visual comparison shows that the CODE framework is more robust at localizing and correctly identifying both known and unknown targets across various complex environments compared to the baseline.}
  \label{fig:vis}
\end{figure*}

\subsection{Ablation Study}

We first conduct an incremental ablation on Surgery to examine how the
three modules interact, followed by removal-based ablations on Aquatic
and Surgery. All experiments in this subsection use OWL-ViT L/14.

\begin{table}[t]
  \caption{Incremental ablation on Surgery Task~1 using OWL-ViT L/14.}
  \label{tab:incremental_ablation}
  \centering
  \begin{tabular}{lcc}
    \toprule
    \textbf{Configuration} & \textbf{U-mAP} & \textbf{K-mAP} \\
    \midrule
    Baseline & 16.2 & 43.0 \\
    + CMJCC & 17.0 & 45.0 \\
    + CMJCC + UGUOE & 22.4 & 42.3 \\
    + CMJCC + UGUOE + DOSCM & 25.7 & 43.9 \\
    \bottomrule
  \end{tabular}
\end{table}

\begin{table}[t]
  \caption{Removal-based ablation on Aquatic and Surgery Task~1 using
  OWL-ViT L/14.}
  \label{tab:ablation}
  \centering
  \begin{tabular}{l|cc|cc}
    \toprule
    \textbf{Configuration} &
    \multicolumn{2}{c|}{\textbf{Aquatic}} &
    \multicolumn{2}{c}{\textbf{Surgery}} \\
    & \cellcolor{blue!10}U & \cellcolor{orange!10}K
    & \cellcolor{blue!10}U & \cellcolor{orange!10}K \\
    \midrule
    CODE Full & 24.7 & 62.7 & 25.7 & 43.9 \\
    w/o CMJCC & 20.6 & 54.3 & 23.7 & 41.5 \\
    w/o UGUOE & 9.1 & 63.0 & 2.3 & 44.2 \\
    w/o DOSCM & 8.3 & 62.0 & 22.4 & 42.3 \\
    \bottomrule
  \end{tabular}
\end{table}

Table~\ref{tab:incremental_ablation} shows how the three modules
progressively contribute to the final performance. Adding CMJCC to the
baseline improves U-/K-mAP from $16.2/43.0$ to $17.0/45.0$,
confirming that visual prototypes strengthen known-class
discrimination. UGUOE then raises U-mAP from $17.0$ to $22.4$,
although K-mAP decreases to $42.3$, reflecting its primary role in
activating potential unknown objects. Finally, DOSCM further increases
U-mAP to $25.7$ and recovers K-mAP to $43.9$, demonstrating that
margin-aware suppression complements uncertainty-guided enhancement.

The removal results in Table~\ref{tab:ablation} provide consistent
evidence. Without CMJCC, U-/K-mAP decreases from $24.7/62.7$ to
$20.6/54.3$ on Aquatic and from $25.7/43.9$ to $23.7/41.5$ on
Surgery. Removing UGUOE causes the largest U-mAP degradation, reaching
only $9.1$ on Aquatic and $2.3$ on Surgery, while K-mAP slightly
increases because fewer ambiguous candidates are activated as unknown.
Removing DOSCM also reduces U-mAP to $8.3$ and $22.4$, respectively,
and lowers K-mAP in both datasets. Together, the two ablation protocols
confirm that CMJCC, UGUOE, and DOSCM address complementary stages of
known calibration, unknown activation, and boundary suppression.

\begin{table}[t]
  \caption{Sample-size sensitivity of visual prototypes using OWL-ViT L/14. Values are U-/K-mAP on Task~1.}
  \label{tab:fewshot_prototypes}
  \centering
  \begin{tabular}{lcccc}
    \toprule
    \textbf{Dataset} &
    \textbf{10-shot} &
    \textbf{50-shot} &
    \textbf{75-shot} &
    \textbf{100-shot} \\
    \midrule
    Aquatic & 18.0/35.3 & 21.0/62.5 & 20.5/63.0 & 24.7/62.7 \\
    Surgery & 13.5/39.8 & 26.1/42.2 & 27.4/43.5 & 25.7/43.9 \\
    \bottomrule
  \end{tabular}
\end{table}

\subsection{Alternative Uncertainty and OOD Scores}

We compare CODE with standard alternatives on Surgery Task~1 using
L/14. DOSCM achieves $25.7$ U-mAP, compared with $21.2$ for
Energy~\cite{liu2020energy}, $2.8$ for
MaxLogit~\cite{hendrycks2019scaling}, and $1.0$ for global
entropy~\cite{chan2021entropy}. UGUOE also obtains $25.7$ U-mAP,
whereas replacing visual-response variance with known-logit entropy
yields $17.0$. Thus, local visual dispersion and margin-aware
suppression better preserve ambiguous unknown objects than the tested
global logit statistics.

\subsection{Prototype Sensitivity and Efficiency}

Table~\ref{tab:fewshot_prototypes} shows that prototype estimation is
sensitive under extreme scarcity, particularly at 10 shots. With 50 or
more samples, K-mAP becomes substantially more stable, while Aquatic
U-mAP reaches $24.7$ at 100 shots. This experiment isolates sample-size
sensitivity rather than a complete long-tailed or noisy-label setting.

For $N\in\{5,15,25,50\}$ retained attributes on Aquatic, CODE obtains
U-/K-mAP values of $15.5/59.0$, $18.9/62.9$, $24.7/62.7$, and
$21.1/62.5$, respectively. K-mAP is stable once $N>15$, while U-mAP is
highest at $N=25$; we therefore use $N=25$ as the accuracy--efficiency
balance. The additional prototype matching has complexity
$\mathcal{O}(NKD)$, accounts for less than $0.1\%$ of the transformer
backbone FLOPs, and increases latency by less than $5\%$ on a single
RTX~4090.

\subsection{Qualitative Results and Visualization}

Figure~\ref{fig:vis} compares CODE with PASS across the five RWD
datasets. In Task~1, CODE recovers the camouflaged
\textit{stingray} in Aquatic and the blurred \textit{scalpel} in
Surgery, supporting the ability of UGUOE to activate difficult unknown
objects. In Task~2, CODE retrieves the missed
\textit{basketballcourt} in Aerial and more fine-grained targets in
Medical, illustrating the benefit of visual-prototype calibration.
Game remains challenging: although CODE recalls more foreground
instances, highly overlapping representations can still cause category
confusion. Overall, the examples support the quantitative improvements
in both unknown activation and known-class calibration.

\section{Conclusion}

We presented \textbf{CODE}, a unified inference-time framework for
attribute-driven open-world object detection. CMJCC uses known-class
visual prototypes to calibrate text-driven predictions, UGUOE enhances
potential unknowns through local visual uncertainty, and DOSCM protects
ambiguous boundary instances through margin-aware suppression.
Together, these components improve unknown-object discovery and
known-class recognition across diverse domains and backbone scales,
highlighting the benefit of combining direct visual references with
uncertainty-aware inference.

\paragraph{Limitations.}
CODE relies on dataset-specific visual prototypes, whose
representativeness may decrease under extreme data scarcity,
long-tailed or noisy annotations, and cross-domain distribution shift.
Our few-shot study isolates sample scarcity but does not cover all
these conditions. Highly overlapping fine-grained categories, as
observed in Game and Surgery, also remain challenging.

\begin{acks}
This work was supported by the Joint Funds of the National Natural Science Foundation of China (No. U2441206).
\end{acks}

\bibliographystyle{ACM-Reference-Format}
\balance
\bibliography{reference}

@String{Computer = "{IEEE} Computer" }

@String{Springer = "Springer-Verlag" }

@article{kuo2022f,
  title={F-vlm: Open-vocabulary object detection upon frozen vision and language models},
  author={Kuo, Weicheng and Cui, Yin and Gu, Xiuye and Piergiovanni, AJ and Angelova, Anelia},
  journal={arXiv preprint arXiv:2209.15639},
  year={2022}
}

@inproceedings{xi2024ktcn,
  title={KTCN: Enhancing Open-World Object Detection with Knowledge Transfer and Class-Awareness Neutralization.},
  author={Xi, Xing and Huang, Yangyang and Lin, Jinhao and Luo, Ronghua},
  booktitle={IJCAI},
  pages={1462--1470},
  year={2024}
}

@inproceedings{joseph2021towards,
  title={Towards open world object detection},
  author={Joseph, KJ and Khan, Salman and Khan, Fahad Shahbaz and Balasubramanian, Vineeth N},
  booktitle={Proceedings of the IEEE/CVF conference on computer vision and pattern recognition},
  pages={5830--5840},
  year={2021}
}

@inproceedings{gupta2022ow,
  title={Ow-detr: Open-world detection transformer},
  author={Gupta, Akshita and Narayan, Sanath and Joseph, KJ and Khan, Salman and Khan, Fahad Shahbaz and Shah, Mubarak},
  booktitle={Proceedings of the IEEE/CVF conference on computer vision and pattern recognition},
  pages={9235--9244},
  year={2022}
}

@inproceedings{zohar2023prob,
  title={Prob: Probabilistic objectness for open world object detection},
  author={Zohar, Orr and Wang, Kuan-Chieh and Yeung, Serena},
  booktitle={Proceedings of the IEEE/CVF conference on computer vision and pattern recognition},
  pages={11444--11453},
  year={2023}
}

@article{zohar2023open,
  title={Open world object detection in the era of foundation models},
  author={Zohar, Orr and Lozano, Alejandro and Goel, Shelly and Yeung, Serena and Wang, Kuan-Chieh},
  journal={arXiv preprint arXiv:2312.05745},
  year={2023}
}

@article{xi2024umb,
  title={Umb: Understanding model behavior for open-world object detection},
  author={Xi, Xing and Huang, Yangyang and Zhong, Zhijie and Luo, Ronghua},
  journal={Advances in Neural Information Processing Systems},
  volume={37},
  pages={74233--74261},
  year={2024}
}

@inproceedings{yang2025detecting,
  title={Detecting Open World Objects via Partial Attribute Assignment},
  author={Yang, Muli and Goenawan, Gabriel James and Qin, Huaiyuan and Han, Kai and Peng, Xi and Yang, Yanhua and Zhu, Hongyuan},
  booktitle={Proceedings of the Computer Vision and Pattern Recognition Conference},
  pages={20318--20328},
  year={2025}
}

@article{gu2021open,
  title={Open-vocabulary object detection via vision and language knowledge distillation},
  author={Gu, Xiuye and Lin, Tsung-Yi and Kuo, Weicheng and Cui, Yin},
  journal={arXiv preprint arXiv:2104.13921},
  year={2021}
}

@inproceedings{zhong2022regionclip,
  title={Regionclip: Region-based language-image pretraining},
  author={Zhong, Yiwu and Yang, Jianwei and Zhang, Pengchuan and Li, Chunyuan and Codella, Noel and Li, Liunian Harold and Zhou, Luowei and Dai, Xiyang and Yuan, Lu and Li, Yin and others},
  booktitle={Proceedings of the IEEE/CVF conference on computer vision and pattern recognition},
  pages={16793--16803},
  year={2022}
}

@inproceedings{minderer2022simple,
  title={Simple open-vocabulary object detection},
  author={Minderer, Matthias and Gritsenko, Alexey and Stone, Austin and Neumann, Maxim and Weissenborn, Dirk and Dosovitskiy, Alexey and Mahendran, Aravindh and Arnab, Anurag and Dehghani, Mostafa and Shen, Zhuoran and others},
  booktitle={European conference on computer vision},
  pages={728--755},
  year={2022},
  organization={Springer}
}

@inproceedings{zareian2021open,
  title={Open-vocabulary object detection using captions},
  author={Zareian, Alireza and Rosa, Kevin Dela and Hu, Derek Hao and Chang, Shih-Fu},
  booktitle={Proceedings of the IEEE/CVF conference on computer vision and pattern recognition},
  pages={14393--14402},
  year={2021}
}

@article{zhao2023revisiting,
  title={Revisiting open world object detection},
  author={Zhao, Xiaowei and Ma, Yuqing and Wang, Duorui and Shen, Yifan and Qiao, Yixuan and Liu, Xianglong},
  journal={IEEE transactions on circuits and systems for video technology},
  volume={34},
  number={5},
  pages={3496--3509},
  year={2023},
  publisher={IEEE}
}

@article{li2024open,
  title={Open world object detection: A survey},
  author={Li, Yiming and Wang, Yi and Wang, Wenqian and Lin, Dan and Li, Bingbing and Yap, Kim-Hui},
  journal={IEEE Transactions on Circuits and Systems for Video Technology},
  volume={35},
  number={2},
  pages={988--1008},
  year={2024},
  publisher={IEEE}
}

@inproceedings{yao2023detclipv2,
  title={Detclipv2: Scalable open-vocabulary object detection pre-training via word-region alignment},
  author={Yao, Lewei and Han, Jianhua and Liang, Xiaodan and Xu, Dan and Zhang, Wei and Li, Zhenguo and Xu, Hang},
  booktitle={Proceedings of the IEEE/CVF Conference on Computer Vision and Pattern Recognition},
  pages={23497--23506},
  year={2023}
}

@inproceedings{maaz2022class,
  title={Class-agnostic object detection with multi-modal transformer},
  author={Maaz, Muhammad and Rasheed, Hanoona and Khan, Salman and Khan, Fahad Shahbaz and Anwer, Rao Muhammad and Yang, Ming-Hsuan},
  booktitle={European conference on computer vision},
  pages={512--531},
  year={2022},
  organization={Springer}
}

@inproceedings{radford2021learning,
  title={Learning transferable visual models from natural language supervision},
  author={Radford, Alec and Kim, Jong Wook and Hallacy, Chris and Ramesh, Aditya and Goh, Gabriel and Agarwal, Sandhini and Sastry, Girish and Askell, Amanda and Mishkin, Pamela and Clark, Jack and others},
  booktitle={International conference on machine learning},
  pages={8748--8763},
  year={2021},
  organization={PmLR}
}

@article{koner2021oodformer,
  title={Oodformer: Out-of-distribution detection transformer},
  author={Koner, Rajat and Sinhamahapatra, Poulami and Roscher, Karsten and G{\"u}nnemann, Stephan and Tresp, Volker},
  journal={arXiv preprint arXiv:2107.08976},
  year={2021}
}

@article{hendrycks2016baseline,
  title={A baseline for detecting misclassified and out-of-distribution examples in neural networks},
  author={Hendrycks, Dan and Gimpel, Kevin},
  journal={arXiv preprint arXiv:1610.02136},
  year={2016}
}

@article{liu2020energy,
  title={Energy-based out-of-distribution detection},
  author={Liu, Weitang and Wang, Xiaoyun and Owens, John and Li, Yixuan},
  journal={Advances in neural information processing systems},
  volume={33},
  pages={21464--21475},
  year={2020}
}

@article{hendrycks2019scaling,
  title={Scaling out-of-distribution detection for real-world settings},
  author={Hendrycks, Dan and Basart, Steven and Mazeika, Mantas and Zou, Andy and Kwon, Joe and Mostajabi, Mohammadreza and Steinhardt, Jacob and Song, Dawn},
  journal={arXiv preprint arXiv:1911.11132},
  year={2019}
}

@article{liang2017enhancing,
  title={Enhancing the reliability of out-of-distribution image detection in neural networks},
  author={Liang, Shiyu and Li, Yixuan and Srikant, Rayadurgam},
  journal={arXiv preprint arXiv:1706.02690},
  year={2017}
}

@article{sun2021react,
  title={React: Out-of-distribution detection with rectified activations},
  author={Sun, Yiyou and Guo, Chuan and Li, Yixuan},
  journal={Advances in neural information processing systems},
  volume={34},
  pages={144--157},
  year={2021}
}

@article{huang2021importance,
  title={On the importance of gradients for detecting distributional shifts in the wild},
  author={Huang, Rui and Geng, Andrew and Li, Yixuan},
  journal={Advances in Neural Information Processing Systems},
  volume={34},
  pages={677--689},
  year={2021}
}

@article{ming2022delving,
  title={Delving into out-of-distribution detection with vision-language representations},
  author={Ming, Yifei and Cai, Ziyang and Gu, Jiuxiang and Sun, Yiyou and Li, Wei and Li, Yixuan},
  journal={Advances in neural information processing systems},
  volume={35},
  pages={35087--35102},
  year={2022}
}

@inproceedings{wang2023clipn,
  title={Clipn for zero-shot ood detection: Teaching clip to say no},
  author={Wang, Hualiang and Li, Yi and Yao, Huifeng and Li, Xiaomeng},
  booktitle={Proceedings of the IEEE/CVF International Conference on Computer Vision},
  pages={1802--1812},
  year={2023}
}

@article{ming2024does,
  title={How does fine-tuning impact out-of-distribution detection for vision-language models?},
  author={Ming, Yifei and Li, Yixuan},
  journal={International Journal of Computer Vision},
  volume={132},
  number={2},
  pages={596--609},
  year={2024},
  publisher={Springer}
}

@inproceedings{zhang2024vision,
  title={Vision-language dual-pattern matching for out-of-distribution detection},
  author={Zhang, Zihan and Xu, Zhuo and Xiang, Xiang},
  booktitle={European Conference on Computer Vision},
  pages={273--291},
  year={2024},
  organization={Springer}
}

@inproceedings{esmaeilpour2022zero,
  title={Zero-shot out-of-distribution detection based on the pre-trained model clip},
  author={Esmaeilpour, Sepideh and Liu, Bing and Robertson, Eric and Shu, Lei},
  booktitle={Proceedings of the AAAI conference on artificial intelligence},
  volume={36},
  number={6},
  pages={6568--6576},
  year={2022}
}

@inproceedings{zhou2022conditional,
  title={Conditional prompt learning for vision-language models},
  author={Zhou, Kaiyang and Yang, Jingkang and Loy, Chen Change and Liu, Ziwei},
  booktitle={Proceedings of the IEEE/CVF conference on computer vision and pattern recognition},
  pages={16816--16825},
  year={2022}
}

@article{fort2021exploring,
  title={Exploring the limits of out-of-distribution detection},
  author={Fort, Stanislav and Ren, Jie and Lakshminarayanan, Balaji},
  journal={Advances in neural information processing systems},
  volume={34},
  pages={7068--7081},
  year={2021}
}

@article{ciaglia2022roboflow,
  title={Roboflow 100: A rich, multi-domain object detection benchmark},
  author={Ciaglia, Floriana and Zuppichini, Francesco Saverio and Guerrie, Paul and McQuade, Mark and Solawetz, Jacob},
  journal={arXiv preprint arXiv:2211.13523},
  year={2022}
}

@article{li2020object,
  title={Object detection in optical remote sensing images: A survey and a new benchmark},
  author={Li, Ke and Wan, Gang and Cheng, Gong and Meng, Liqiu and Han, Junwei},
  journal={ISPRS journal of photogrammetry and remote sensing},
  volume={159},
  pages={296--307},
  year={2020},
  publisher={Elsevier}
}

@article{bouget2015detecting,
  title={Detecting surgical tools by modelling local appearance and global shape},
  author={Bouget, David and Benenson, Rodrigo and Omran, Mohamed and Riffaud, Laurent and Schiele, Bernt and Jannin, Pierre},
  journal={IEEE transactions on medical imaging},
  volume={34},
  number={12},
  pages={2603--2617},
  year={2015},
  publisher={IEEE}
}

@article{menon2022visual,
  title={Visual classification via description from large language models},
  author={Menon, Sachit and Vondrick, Carl},
  journal={arXiv preprint arXiv:2210.07183},
  year={2022}
}

@article{zohar2023lovm,
  title={Lovm: Language-only vision model selection},
  author={Zohar, Orr and Huang, Shih-Cheng and Wang, Kuan-Chieh and Yeung, Serena},
  journal={Advances in Neural Information Processing Systems},
  volume={36},
  pages={33120--33132},
  year={2023}
}

@inproceedings{lin2024generative,
  title={Generative region-language pretraining for open-ended object detection},
  author={Lin, Chuang and Jiang, Yi and Qu, Lizhen and Yuan, Zehuan and Cai, Jianfei},
  booktitle={2024 IEEE/CVF Conference on Computer Vision and Pattern Recognition (CVPR)},
  pages={13958--13968},
  year={2024},
  organization={IEEE}
}

@article{cao2025open,
  title={Open-Det: An efficient learning framework for open-ended detection},
  author={Cao, Guiping and Wang, Tao and Huang, Wenjian and Lan, Xiangyuan and Zhang, Jianguo and Jiang, Dongmei},
  journal={arXiv preprint arXiv:2505.20639},
  year={2025}
}

@inproceedings{liu2025detr,
  title={UN-DETR: Promoting objectness learning via joint supervision for unknown object detection},
  author={Liu, Haomiao and Xu, Hao and Yue, Chuhuai and Ma, Bo},
  booktitle={Proceedings of the AAAI Conference on Artificial Intelligence},
  volume={39},
  number={5},
  pages={5442--5450},
  year={2025}
}

@inproceedings{liu2024uoa,
  title={UOA-RCNN: Detect Anything with Unknown Object Aware RCNN},
  author={Liu, Haomiao and Xu, Hao and Yue, Chuhuai and Ma, Bo},
  booktitle={International Conference on Neural Information Processing},
  pages={63--77},
  year={2024},
  organization={Springer}
}

@article{liu2025adaptive,
  title={Adaptive objectness learning for enhanced unknown object detection: H. Liu et al.},
  author={Liu, Haomiao and Xu, Hao and Yue, Chuhuai and Ma, Bo},
  journal={The Visual Computer},
  volume={41},
  number={10},
  pages={7433--7446},
  year={2025},
  publisher={Springer}
}

@inproceedings{xu2024tgtrack,
  title={TGTrack: Text Modality Autoregression and Generative Template Updating for Visual Object Tracking},
  author={Xu, Hao and Liang, Yiding and Liu, Haomiao and Yue, Chuhuai and Ma, Bo},
  booktitle={International Conference on Neural Information Processing},
  pages={259--273},
  year={2024},
  organization={Springer}
}

@inproceedings{gupta2019lvis,
  title={Lvis: A dataset for large vocabulary instance segmentation},
  author={Gupta, Agrim and Dollar, Piotr and Girshick, Ross},
  booktitle={2019 IEEE/CVF Conference on Computer Vision and Pattern Recognition (CVPR)},
  pages={5351--5359},
  year={2019},
  organization={IEEE}
}

@inproceedings{sun2024exploring,
  title={Exploring orthogonality in open world object detection},
  author={Sun, Zhicheng and Li, Jinghan and Mu, Yadong},
  booktitle={2024 IEEE/CVF Conference on Computer Vision and Pattern Recognition (CVPR)},
  pages={17302--17312},
  year={2024},
  organization={IEEE}
}

@article{heo2026detecting,
  title={Detecting Unknown Objects via Energy-based Separation for Open World Object Detection},
  author={Heo, Jun-Woo and Park, Keonhee and Park, Gyeong-Moon},
  journal={arXiv preprint arXiv:2603.29954},
  year={2026}
}

@inproceedings{chan2021entropy,
  title={Entropy maximization and meta classification for out-of-distribution detection in semantic segmentation},
  author={Chan, Robin and Rottmann, Matthias and Gottschalk, Hanno},
  booktitle={2021 IEEE/CVF International Conference on Computer Vision (ICCV)},
  pages={5108--5117},
  year={2021},
  organization={IEEE}
}

\appendix

\setcounter{equation}{0}
\renewcommand{\theequation}{S\arabic{equation}}

\setcounter{figure}{0}
\renewcommand{\thefigure}{S\arabic{figure}}

\setcounter{table}{0}
\renewcommand{\thetable}{S\arabic{table}}

\section*{Appendix Contents}

\noindent
\hyperref[sec:dataset_details]{A\quad Dataset Details}
\dotfill \pageref{sec:dataset_details}\\
\hspace*{1em}\hyperref[sec:benchmark_composition]{A.1\quad Benchmark Composition}
\dotfill \pageref{sec:benchmark_composition}\\
\hspace*{1em}\hyperref[sec:appendix_stats]{A.2\quad Statistical Summary}
\dotfill \pageref{sec:appendix_stats}\\
\hspace*{1em}\hyperref[sec:task_formulation]{A.3\quad Task Formulation}
\dotfill \pageref{sec:task_formulation}\\
\hspace*{1em}\hyperref[sec:evaluation_tasks]{A.4\quad Evaluation Tasks}
\dotfill \pageref{sec:a4}\\

\noindent
\hyperref[sec:appendix_implementation]{B\quad Implementation Details}
\dotfill \pageref{sec:appendix_implementation}\\
\hspace*{1em}\hyperref[sec:attr_gen]{B.1\quad Attribute Generation and Prompt Templates}
\dotfill \pageref{sec:attr_gen}\\
\hspace*{1em}\hyperref[sec:training_optimization]{B.2\quad Model Training and Optimization}
\dotfill \pageref{sec:training_optimization}\\

\noindent
\hyperref[sec:additional_results]{C\quad Additional Experimental Results and Analysis}
\dotfill \pageref{sec:additional_results}\\
\hspace*{1em}\hyperref[sec:intra_ablations]{C.1\quad Intra-module Ablations}
\dotfill \pageref{sec:intra_ablations}\\
\hspace*{1em}\hyperref[sec:parameter_sensitivity]{C.2\quad Comprehensive Parameter Sensitivity Analysis}
\dotfill \pageref{sec:parameter_sensitivity}\\
\hspace*{1em}\hyperref[sec:visualizations]{C.3\quad Advanced Visualizations and In-depth Analysis}
\dotfill \pageref{sec:visualizations}\\

\noindent
\hyperref[sec:limitations_future_work]{D\quad Limitations and Future Work}
\dotfill \pageref{sec:limitations_future_work}


\section{Dataset Details}
\label{sec:dataset_details}

We evaluate our framework on the Real-World Object Detection (RWD) benchmark, following the experimental protocol established by FOMO \cite{zohar2023open}. The RWD benchmark is designed to assess object detectors' ability to handle known and unknown objects across diverse and challenging domains. 

\subsection{Benchmark Composition}
\label{sec:benchmark_composition}
The benchmark consists of five datasets representing different complex scenarios. Following the RWD setup \cite{zohar2023open}:
\begin{itemize}
    \item \textbf{Aquatic, Game, and Medical:} These datasets are derived from RoboFlow100 \cite{ciaglia2022roboflow}, specifically the aquarium, team-fight-tactics, and xray-rheumatology subsets. They represent underwater environments, synthetic game snapshots with diverse avatars, and hand X-ray images for bone detection, respectively.
    \item \textbf{Aerial:} This dataset is sourced from DIOR \cite{li2020object}, consisting of high-resolution aerial imagery of structures such as stadiums, storage tanks, and ships.
    \item \textbf{Surgery:} Taken from the NeuroSurgicalTools dataset \cite{bouget2015detecting}, these images were captured via neurosurgical microscopes and contain various fine-grained surgical instruments.
\end{itemize}

\subsection{Statistical Summary}
\label{sec:appendix_stats}
Table~\ref{tab:appendix_stats} provides a comprehensive statistical overview of the RWD benchmark. This includes the total image count (training and testing), the class split between known (K) and unknown (U) categories for Task 1, the size of the attribute pool generated via LLMs, and the percentage of class names that exist within the vision-language tokenizer's vocabulary.

\subsection{Task Formulation}
\label{sec:task_formulation}
Open World Object Detection (OWOD) requires a model to simultaneously detect known objects, discover unknown objects, and incrementally learn these novel categories over time. Formally, the OWOD process is divided into a series of subtasks $\mathcal{T} = \{T_1, T_2, \dots, T_{|\mathcal{T}|}\}$. At a specific task stage $t$, the model is trained on a set of known object classes denoted as $K^t = \{O^t_1, O^t_2, \dots, O^t_{|K^t|}\}$. During evaluation, the model is expected to detect all categories it has encountered so far (i.e., $K^t$), as well as discover unlabeled but interesting unknown categories, denoted as $U^t$. 

After discovering the unknown object classes, the model can be updated with the knowledge of these new classes using annotations from an oracle (e.g., a human annotator) in the subsequent task $t+1$. Consequently, the model's known category set expands to include both previously seen classes and the newly introduced ones: $K^{t+1} = K^t \cup U^t$. This readies the model to detect the newly updated known classes alongside additional unknown classes for the next task cycle. This formulation describes the general OWOD paradigm. In the RWD
evaluation used in this work, however, Task~2 follows the benchmark
protocol described in Sec.~4.1.1 of the main paper: the model is jointly
retrained using all samples available at that stage. Therefore, our
Task~2 results should be interpreted as evaluation over an expanded
category set rather than as strictly sequential incremental learning.

\begin{table}[h]
\centering
\caption{Comprehensive summary of the RWD benchmark datasets. We report the total images, class distribution for Task 1, attribute pool size, and tokenizer vocabulary coverage (\%).}
\label{tab:appendix_stats}
\resizebox{\columnwidth}{!}{
\begin{tabular}{lcccc}
\toprule
\textbf{Dataset} & \textbf{Total Images} & \textbf{Classes (K+U)} & \textbf{Attributes} & \textbf{\% in Tokenizer} \\ \midrule
Aquatic & 637   & 7 (4+3)   & 385   & 100.0 \\
Aerial  & 10,000 & 20 (10+10) & 1,229  & 55.0  \\
Game    & 1,575  & 59 (30+29) & 390   & 35.6  \\
Medical & 182   & 12 (6+6)  & 390   & 15.4  \\
Surgery & 1,829  & 13 (6+7)  & 808   & 7.69  \\ \bottomrule
\end{tabular}
}
\end{table}

\subsection{Evaluation Tasks}
\label{sec:a4}
Following the scarcity-based split motivated by long-tailed recognition 
, each dataset is divided into two subsets: the 50\% most common classes (Known) and the 50\% least common classes (Unknown). Within our specific two-stage evaluation protocol:
\begin{itemize}
    \item \textbf{Task 1:} Detectors are trained on known classes ($K^1$) and must generalize to detect novel, unlabeled unknown objects. All categories in the test set that belong to the subsequent task are treated as unknown classes ($U^1$).
   \item \textbf{Task 2:} Previously unknown classes are revealed and
added to the expanded known-category set
($K^2 = K^1 \cup U^1$). Following the RWD protocol used in the main
paper, the model is jointly retrained using all samples available at
this stage. The categories from Task~1 are evaluated as Previously
Known (PK), while the newly introduced categories are evaluated as
Currently Known (CK).
\end{itemize}

\section{Implementation Details}
\label{sec:appendix_implementation}

\subsection{Attribute Generation and Prompt Templates}
\label{sec:attr_gen}

To provide a fine-grained semantic foundation for attribute reasoning, we utilize LLMs to decompose abstract category labels into class-agnostic attributes, following the protocols established by previous works \cite{zohar2023open, xi2024umb, yang2025detecting}. This process allows the model to establish an initial semantic baseline for discovering unknown objects.

\subsubsection{LLM-based Decomposition}
We prompt GPT-3.5 to generate descriptive characteristics for each known class within the RWD benchmark. The following prompt template is employed to ensure that the generated attributes remain visually or functionally discriminative:
\begin{quote}
    \textit{``I am using a language-vision model to identify \{Category\}. List the \{Type\} attributes of \{Category\}, which will be used for detection.''}
\end{quote}
where \texttt{\{Category\}} denotes the specific class name and \texttt{\{Type\}} refers to one of the ten semantic dimensions. This methodology leverages the extensive internal knowledge of LLMs to facilitate the reasoning of unknown objects that share latent characteristics with known classes.

\subsubsection{Attribute Taxonomy}
To ensure a comprehensive characterization of objects across diverse and complex domains, the generated attributes are categorized into ten distinct dimensions:
\begin{itemize}
    \item \textbf{Shape:} e.g., flat disc-like, pointed tips, knobby.
    \item \textbf{Color:} e.g., silver color, turquoise, yellow.
    \item \textbf{Texture:} e.g., matte skin, ridged, fissured.
    \item \textbf{Size:} e.g., asymmetric, large, proportionality.
    \item \textbf{Context:} e.g., underwater, surgical bone mallet, wrist bones.
    \item \textbf{Features:} e.g., rounded snout, tension adjustment, cortical bone.
    \item \textbf{Appearance:} e.g., shiny, presence of parking lots, visible joints.
    \item \textbf{Behavior:} e.g., swimming, stabilizing, extension.
    \item \textbf{Environment:} e.g., artificial reef, surgical tools, sparse.
    \item \textbf{Material:} e.g., collagen, bony, hydroxyapatite.
\end{itemize}

\subsubsection{Encoding for Attribute Reasoning}
To align the raw attribute descriptions with the pre-training distribution of the foundation model, we wrap each attribute into a structured descriptive sentence \cite{menon2022visual, zohar2023lovm}:
\begin{quote}
    \textit{``Object which (is/has/etc) \{Type\} is \{Attribute\}.''}
\end{quote}
These formatted descriptions are then encoded into high-dimensional embeddings using the frozen text encoder of OWL-ViT \cite{minderer2022simple}. This procedure constructs the necessary semantic space for the model's attribute reasoning, providing a structured textual prior that enables the detection of potential unknown objects based on their constituent properties.

\subsection{Model Training and Optimization}
\label{sec:training_optimization}

\subsubsection{Training and Loss Functions}
The training process of CODE is supervised by a composite objective function (Eq. 1 in the main text), formulated as:
\begin{equation}
    \mathcal{L} = \mathcal{L}_{CE} + \lambda_{POT} \mathcal{L}_{POT} + \lambda_{MSE} \mathcal{L}_{MSE} + \lambda_{L1} \mathcal{L}_{L1}
\end{equation}
where the empirical trade-off coefficients are set to $\lambda_{POT} = 5$, $\lambda_{MSE} = 0.1$, and $\lambda_{L1} = 0.001$ across all experiments. Each component is detailed as follows:

\begin{itemize}
    \item \textbf{Partial Optimal Transport Loss ($\mathcal{L}_{POT}$):} 
    Following the assignment strategy in \cite{yang2025detecting}, conventional Optimal Transport (OT) assumes identical total probability mass between two distributions ($\|V\|_1 = \|A\|_1$). However, this fails in our scenario because only a small subset of the large generated attribute pool $A$ aligns with the visual object embeddings $V$. To address this, $\mathcal{L}_{POT}$ relaxes the equal mass assumption, transporting only the in-distribution mass between $V$ and $A$ while explicitly filtering out redundant OOD attributes. This is formulated with asymmetric marginal constraints:
    \begin{equation}
        \Pi(V,A) \triangleq \{T \in \mathbb{R}_{+}^{M \times N} \mid T \mathbf{1}_N = V, T^\top \mathbf{1}_M \leqslant A \}
    \end{equation}
    Unlike traditional POT, this introduces an equality constraint on the visual distribution $V$ and an inequality constraint on $A$. The loss is computed as $d_{POT,\epsilon}(V,A;C) \triangleq \min_{T \in \Pi(V,A)} \langle T,C \rangle_F - \epsilon h(T)$, where $h(T)$ is an entropic constraint added to accelerate convergence.

    \item \textbf{Cross-Entropy Loss ($\mathcal{L}_{CE}$):} 
    After acquiring the selected attributes $A'$ via POT, we learn a mapping matrix $\mathbf{W} = [w_1, w_2, ..., w_K]$ to assign these attributes to $K$ known object classes. The prediction probability $p(O_k|v)$ is computed via a softmax function over the similarity between visual embeddings and the reweighted attributes. The standard cross-entropy loss is then applied to supervise the optimization of both the mapping matrix and the selected attributes:
    \begin{equation}
        \mathcal{L}_{CE} = -\frac{1}{M} \sum_{m=1}^{M} \sum_{k=1}^{K} l_{m,k} \log p(O_k|v_m)
    \end{equation}
    where $l_{m,k}$ is the one-hot label vector for the visual patch $v_m$, and $M$ is the number of known image patches.

    \item \textbf{Mean Squared Error Loss ($\mathcal{L}_{MSE}$):} 
    Applied to align the text-derived attributes with the average vision-derived class embeddings. This ensures stable feature refinement and cross-modal semantic consistency under limited bounding-box supervision.

    \item \textbf{L1 Regularization ($\mathcal{L}_{L1}$):} 
    Applied to the attribute mapping weights $\mathbf{W}$ (i.e., $\|\mathbf{W}\|_1$) to induce sparsity. This encourages the model to selectively activate only the most discriminative attributes rather than redundantly using the entire pool.
\end{itemize}

\subsubsection{Inference Pipeline}
The complete inference pipeline of our proposed framework is summarized in Algorithm~\ref{alg:CMJCC}. By sequentially executing CMJCC, UGUOE, and DOSCM, our method elegantly resolves the semantic ambiguity of known objects and the over-suppression of unknown targets within a unified, end-to-end inference pass. Specifically, the CMJCC module first provides visual-to-visual calibration, followed by UGUOE for boosting potential unknown regions, and finally DOSCM to dynamically protect boundary targets.

\begin{algorithm}[tb]
    \caption{Inference Procedure of CODE for OWOD}
    \label{alg:CMJCC}
    \KwIn{Candidate visual embedding $\boldsymbol{v}_i$, visual prototypes
$\boldsymbol{V}_{base}$, text-driven known logits $Logit_{K,i}$,
text-driven unknown logit $Logit_{U,i}$, task relevance score $p_{ID}$.}
    \KwOut{Calibrated known probability
$P_K^{calib}(\boldsymbol{v}_i)$, final unknown probability
$P_U^{final}(\boldsymbol{v}_i)$.}
    
    \tcp*[h]{Cross-Modal Joint Confidence Calibration (CMJCC)}\\
    Compute visual similarities:
$Sim_{vis,i} \leftarrow
\boldsymbol{v}_i \cdot \boldsymbol{V}_{base}^{\top}$\;

Calculate candidate-specific threshold:
$\tau_i \leftarrow
\max\left(
\frac{1}{K}\sum_{k=1}^{K}[Sim_{vis,i}]_k + m,\,
\tau_{min}
\right)$\;

Compute visual boost:
$Logit_{boost,i} \leftarrow
\text{ReLU}(Sim_{vis,i}-\tau_i)$\;

Calibrate known logits:
$Logit_{K,i}^{calib} \leftarrow
Logit_{K,i} +
\alpha \cdot S_{scale} \cdot Logit_{boost,i}$\;

Calculate calibrated known probabilities:
$P_K^{calib}(\boldsymbol{v}_i)
\leftarrow
\sigma(Logit_{K,i}^{calib})$\;
\tcp*[h]{For brevity, the candidate index $i$ is omitted below.}
    
    \vspace{1mm}
    \tcp*[h]{Uncertainty-Guided Universal Objectness Enhancement (UGUOE)}\\
    Extract Top-$K_u$ local similarities $S_{local}$ from $Sim_{vis}$\;
    Compute base unknown response $\mu_{unk}$ and local variance
$\sigma^{2}(S_{local})$ following the definitions in Sec.~3.3\;
    Calculate hesitation weight $W_{unc} \leftarrow \exp(-\gamma \cdot \sigma^{2}(S_{local}))$ and clamp to $[w_{min}, w_{max}]$\;
    Calculate universal objectness: $Obj_{unk} \leftarrow \text{ReLU}(\mu_{unk} - \tau_{unk})$\;
    Enhance unknown logit: $Logit_{U}^{enh} \leftarrow Logit_{U} + \beta \cdot S_{scale} \cdot W_{unc} \cdot Obj_{unk}$\;
    
    \vspace{1mm}
    \tcp*[h]{Dynamic Outlier Suppression via Confidence Margin (DOSCM)}\\
    Extract Top-1 and Top-2 probabilities from
$P_K^{calib}(\boldsymbol{v}_i)$: $P_{top1}, P_{top2}$\;
    Calculate confidence margin: $\Delta P \leftarrow P_{top1} - P_{top2}$\;
    Compute effective suppression penalty: $MCM_{eff} \leftarrow P_{top1} \cdot \Delta P$\;
    Compute dynamic OOD score: $p_{OOD}^{dyn} \leftarrow 1 - MCM_{eff}$\;
    Calculate final unknown probability: $P_U^{final}(\boldsymbol{v}_i) \leftarrow \sigma(Logit_{U}^{enh}) \cdot p_{ID} \cdot p_{OOD}^{dyn}$\;
    
    \vspace{1mm}
    \Return $P_K^{calib}(\boldsymbol{v}_i)$,
$P_U^{final}(\boldsymbol{v}_i)$\;
\end{algorithm}

\subsubsection{Hyperparameters, Environment, and Strategy}
All experiments are conducted on a server equipped with NVIDIA RTX 4090 GPUs. We use the AdamW optimizer with a weight decay of $1 \times 10^{-4}$. The text encoder of the pre-trained OWL-ViT remains frozen during training to preserve its zero-shot generalization capabilities. For the B/16 backbone, the batch size is set to 10 with input images resized to $768 \times 768$. For the L/14 backbone, we use a batch size of 1 and resize images to $840 \times 840$. During both training and inference, the maximum number of predicted bounding boxes per image is fixed at 100.

For few-shot training, we follow \cite{zohar2023open, yang2025detecting} to feed an image and its corresponding ground truth bounding box into the pre-trained OWL-ViT \cite{minderer2022simple} model to generate predicted bounding boxes and class embeddings. The class embeddings are filtered based on their associated bounding boxes, ensuring that only those with an intersection over union (IoU) of at least 0.8 with the ground-truth object are retained.  Prompt ensembling is also used to produce the final attribute embeddings by averaging the text embeddings obtained from the 7 most effective CLIP prompt templates \cite{radford2021learning, minderer2022simple}.

\textbf{Additional prototype extraction detail.} As described in Sec.~3.2 of the main paper, visual prototypes are
constructed only from labeled data available at the corresponding
stage. In implementation, the image and its ground-truth bounding
boxes are passed through the frozen OWL-ViT model, and region
embeddings associated with predicted boxes whose IoU with the
corresponding ground-truth box is at least $0.8$ are retained for
class-wise prototype estimation. The caching and Task~1/Task~2
prototype-bank protocol follows Sec.~3.2.

To adapt to the diverse data distributions in the RWD benchmark, we
select the optimization and inference hyperparameters through grid
search. For training, the learning rate is searched over
$LR \in [1 \times 10^{-3}, 5 \times 10^{-3}, 5 \times 10^{-4}]$,
and the maximum number of iterations over
$T_{max} \in [1, 10, 100]$. The remaining CODE hyperparameters are
selected in the same grid-search manner, with the final settings
reported in Sec.~4.1.3 of the main paper.

For the quantitative comparison in the main paper, results of previous
methods are taken from their original publications, while CODE is
evaluated using our final configuration. Reproduced baseline
predictions, where used in the qualitative visualizations of this
appendix, are used only for visualization and are not substituted for
the published quantitative results.

\textbf{Baseline Implementations:} To comprehensively evaluate the effectiveness of our CODE framework, we compare it against state-of-the-art OWOD methods alongside several strong baselines adapted from Open-Vocabulary Object Detection (OVOD) using the OWL-ViT foundation model (Tab. 1 in the main text). These baselines are configured as follows:
\begin{itemize}
    \item \textbf{BASE-ZS:} Operates in a pure zero-shot setting by using a generic prompt (e.g., ``object'' or ``a photo of an object'') to detect unknown targets \cite{maaz2022class}.
    \item \textbf{BASE-ZS+IN:} Utilizes the comprehensive ImageNet vocabulary as proposals for unknown objects, explicitly removing the names of the currently known classes to prevent overlap.
    \item \textbf{BASE-ZS+LLM:} Leverages an LLM to predict potential unknown object categories based on the semantic context of the given known classes. These generated names are then used as text queries for zero-shot detection.
    \item \textbf{BASE-ZS+GT:} Employs the actual ground-truth class names for the unknown objects. This configuration acts as the theoretical upper bound for text-conditioned (zero-shot) OVOD approaches, as it assumes oracle access to the exact identities of the unknown targets.
    \item \textbf{BASE-FS:} A few-shot baseline that receives the same level of visual supervision as our method. It uses image exemplars to extract vision-derived object embeddings, which are averaged per class to generate visual class prototypes. It then applies a generic text prompt to identify unknown objects. Because text-based embeddings inherently exhibit different cosine similarity scales compared to image-derived counterparts, the predictions for known and unknown objects are ranked and selected separately.
\end{itemize}

\section{Additional Experimental Results and Analysis}
\label{sec:additional_results}

The main paper already reports the incremental module ablation
(Table~2), alternative uncertainty and OOD-score comparisons
(Sec.~4.4), prototype sample-size and attribute-number sensitivity
(Table~4 and Sec.~4.5), computational overhead, and generalization to
standard benchmarks (Sec.~4.2). To avoid duplicating these results,
this appendix focuses on complementary intra-module ablations,
controlled parameter sweeps, and additional visual analyses.

\subsection{Intra-module Ablations}
\label{sec:intra_ablations}

To further validate the precise architectural design within our proposed framework, we conduct detailed intra-module ablation studies. Here, we investigate the core mechanisms within two critical modules: the dynamic thresholding gate in the \textbf{Cross-Modal Joint Confidence Calibration (CMJCC)} module, and the internal components of the \textbf{Uncertainty-Guided Universal Objectness Enhancement (UGUOE)} module. All experiments are conducted using the L/14 backbone across two representative datasets: Aquatic and Surgery.

\subsubsection{CMJCC Module Ablation}
The CMJCC module explicitly injects global visual prototypes to calibrate text-driven confidences. A core component of this module is the candidate-specific dynamic
thresholding gate
$\tau_i =
\max\left(
\frac{1}{K}\sum_{k=1}^{K}[Sim_{vis,i}]_k + m,\tau_{min}
\right)$, which dictates how the visual similarity $Sim_{vis}$ is transformed into the visual boost $Logit_{boost}$. To isolate the contribution of this dynamic gating, we compare the full CODE framework against two degraded variants under the Task 2 evaluation protocol:

\begin{itemize}
    \item \textbf{w/ Fixed $\tau_{min}$ Only:}
We remove the candidate-specific component
$\frac{1}{K}\sum_{k=1}^{K}[Sim_{vis,i}]_k+m$
and rely solely on the static threshold floor $\tau_{min}$ to gate
the similarities.

\item \textbf{w/o Threshold (Direct $Sim_{vis,i}$):}
We completely remove the subtraction of $\tau_i$ and the ReLU
activation, directly using the raw visual similarity $Sim_{vis,i}$
for enhancement.
\end{itemize}

\begin{table}[htbp]
  \caption{Ablation study of the CMJCC module's dynamic gating mechanism. We report PK-mAP (PK) and CK-mAP (CK) on the Aquatic and Surgery datasets under Task 2 evaluation using the L/14 backbone.}
  \label{tab:ablation_cmjcc}
  \centering
  \small
  \begin{tabular}{l|cc|cc}
    \toprule
    \textbf{Dataset ($\rightarrow$)} & \multicolumn{2}{c|}{\textbf{Aquatic}} & \multicolumn{2}{c}{\textbf{Surgery}} \\
    \midrule
    \textbf{Method Variant} & \cellcolor{green!10}PK & \cellcolor{green!10}CK & \cellcolor{green!10}PK & \cellcolor{green!10}CK \\
    \midrule
    CODE Full (Ours) & \textbf{65.3} & \textbf{59.6} & \textbf{46.3} & \textbf{35.6} \\
    w/ Fixed $\tau_{min}$ Only & 61.7 & 55.6 & 42.3 & 27.6 \\
    w/o Threshold (Direct $Sim_{vis}$) & 59.9 & 49.1 & 42.1 & 25.4 \\
    \bottomrule
  \end{tabular}
\end{table}

\textbf{Analysis:} As shown in Table~\ref{tab:ablation_cmjcc}, the
dynamic gating strategy is crucial for maintaining high performance
under the expanded-category Task~2 setting.
First, replacing the adaptive threshold with a fixed $\tau_{min}$ leads to a significant performance drop, particularly for Currently Known (CK) objects (e.g., dropping from 35.6 to 27.6 on the Surgery dataset). This indicates that a rigid threshold fails to adapt to the
candidate-specific distribution of visual similarities, leading to
either under-activation of valid visual priors or the unintended
inclusion of background noise. As clarified in Sec.~3.2, the dynamic
threshold is computed independently for each candidate and therefore
does not depend on inference-batch composition.

Second, completely removing the gating threshold (Direct $Sim_{vis}$) results in severe performance degradation across all metrics (e.g., CK drops to 49.1 on Aquatic and 25.4 on Surgery). Without the ReLU-based filtering mechanism, background noise and irrelevant local visual responses are indiscriminately injected into the calibration process. This pollutes the original text-driven logits $Logit_{K}$, exacerbating the semantic ambiguity rather than resolving it. Therefore, the proposed adaptive thresholding precisely isolates high-confidence visual priors, ensuring that $Logit_{boost}$ effectively compensates for textual gaps without introducing detrimental noise.

\subsubsection{UGUOE Module Ablation}
The UGUOE module is specifically designed to bridge the gap in unknown object discovery by boosting the responses of potential unknown targets located in the ``semantic vacuum''. To validate its internal mechanisms, we ablate its two core components: the base universal objectness activation $Obj_{unk}$ and the uncertainty-guided weight $W_{unc}$. Experiments are conducted on Task 1 using the L/14 backbone across the Aquatic and Surgery datasets. We compare the full CODE framework with two variants:
\begin{itemize}
    \item \textbf{w/o $Obj_{unk}$ (No Enhancement):} We completely remove the universal objectness injection, reverting the unknown logit to the pure text-driven baseline ($Logit_U^{enh} = Logit_U$).
    \item \textbf{w/o $W_{unc}$ (No Uncertainty Weight):} We remove the variance-based hesitation modeling, treating all regions with high visual similarity equally by setting $W_{unc} = 1$.
\end{itemize}

\begin{table}[htbp]
  \caption{Ablation study of the UGUOE module's internal components. We report U-mAP (U) and K-mAP (K) on the Aquatic and Surgery datasets under Task 1 evaluation.}
  \label{tab:ablation_uguoe}
  \centering
  \small
  \begin{tabular}{l|cc|cc}
    \toprule
    \textbf{Dataset ($\rightarrow$)} & \multicolumn{2}{c|}{\textbf{Aquatic}} & \multicolumn{2}{c}{\textbf{Surgery}} \\
    \midrule
    \textbf{Method Variant} & \cellcolor{blue!10}U & \cellcolor{orange!10}K & \cellcolor{blue!10}U & \cellcolor{orange!10}K \\
    \midrule
    CODE Full (Ours) & \textbf{24.7} & 62.7 & \textbf{25.7} & 43.9 \\
w/o $Obj_{unk}$ (No Enhancement) & 9.1 & \textbf{63.0} & 2.3 & \textbf{44.2} \\
w/o $W_{unc}$ (No Uncertainty Weight) & 20.1 & 62.9 & 23.9 & 43.5 \\
    \bottomrule
  \end{tabular}
\end{table}

\textbf{Analysis:} As detailed in Table~\ref{tab:ablation_uguoe}, both the base objectness enhancement and the uncertainty-guided weighting are indispensable for robust OWOD.

First, removing the universal objectness enhancement entirely
(\textbf{w/o $Obj_{unk}$}) leads to a substantial collapse in the
detection of unknown objects. For instance, the U-mAP on the Surgery
dataset drops from 25.7 to 2.3, while the Aquatic U-mAP drops from
24.7 to 9.1. This severe degradation confirms our hypothesis that
relying solely on a generic, text-derived attribute response is
insufficient to separate ambiguous unknown targets from complex
background noise. Explicitly injecting aggregated local visual
similarities ($\mu_{unk}$) is essential to recall these unlabeled
instances.

Second, discarding the uncertainty modeling
(\textbf{w/o $W_{unc}$}) consistently reduces U-mAP, while the
known-class performance remains largely unchanged. On Aquatic,
U-mAP decreases from 24.7 to 20.1 while K-mAP changes only from
62.7 to 62.9; on Surgery, U-mAP decreases from 25.7 to 23.9 and
K-mAP from 43.9 to 43.5. Without the variance-based penalty ($\sigma^2(S_{local})$), the module indiscriminately boosts any region with high local visual similarity. This inevitably over-boosts clearly defined known objects or textured backgrounds, causing them to interfere with the delicate text-driven prediction space. By formulating $W_{unc}$ as an exponential decay function of local variance, UGUOE correctly identifies the model's ``classification hesitation'' and selectively targets only the ambiguous regions, thus protecting known class precision while steadily improving unknown recall.

\subsection{Comprehensive Parameter Sensitivity Analysis}
\label{sec:parameter_sensitivity}

In this section, we comprehensively investigate the sensitivity of the CODE framework to critical hyper-parameters introduced in two core modules: CMJCC and UGUOE. For the CMJCC module, we analyze the visual gating margin $m$ and the visual prior modulation coefficient $\alpha$ under the Task 2 evaluation protocol. For the UGUOE module, we evaluate the local neighborhood size $K_u$, the uncertainty sensitivity factor $\gamma$, and the unknown activation threshold $\tau_{unk}$ under the Task 1 setting. All evaluations are conducted using the L/14 backbone across the representative Aquatic and Surgery datasets. All reported hyperparameters were selected through grid search.
The tables below retain the values obtained in their corresponding
controlled parameter sweeps and are intended to characterize relative
sensitivity within each sweep; the final full-system performance is
reported separately in the main paper.

\subsubsection{Sensitivity Analysis for CMJCC}
\begin{table}[htbp]
  \caption{Parameter sensitivity analysis of the visual gating margin
$m$ and modulation coefficient $\alpha$. Default parameter values are
highlighted in bold in the \emph{Value} column, while the best metric
values within each parameter sweep are highlighted in bold. PK and CK
indicate Previously Known and Currently Known mAP, respectively.}
  \label{tab:parameter_sensitivity}
  \centering
  \small
  \begin{tabular}{cc|cc|cc}
    \toprule
    \multirow{2}{*}{\textbf{Parameter}} & \multirow{2}{*}{\textbf{Value}} & \multicolumn{2}{c|}{\textbf{Aquatic}} & \multicolumn{2}{c}{\textbf{Surgery}} \\
    & & \cellcolor{green!10}PK & \cellcolor{green!10}CK & \cellcolor{green!10}PK & \cellcolor{green!10}CK \\
    \midrule
    \multirow{5}{*}{\shortstack{Gating\\Margin ($m$)}} 
    & 0 & 50.2 & 41.3 & 40.0 & 23.4 \\
    & 0.1 & 64.0 & 58.0 & 41.8 & 27.6 \\
    & \textbf{0.2} & \textbf{65.3} & 59.6 & 46.3 & 35.6 \\
    & 0.3 & 65.2 & \textbf{59.6} & \textbf{46.4} & \textbf{36.3} \\
    & 0.4 & 62.4 & 58.4 & 43.3 & 32.4 \\
    \midrule
    \multirow{5}{*}{\shortstack{Modulation\\Coef ($\alpha$)}} 
    & 0.1 & 59.5 & 57.6 & 46.4 & 35.4 \\
    & 0.25 & 64.4 & \textbf{60.8} & \textbf{46.8} & \textbf{36.5} \\
    & \textbf{0.5} & \textbf{65.3} & 59.6 & 46.3 & 35.6 \\
    & 1.0 & 63.2 & 55.5 & 45.2 & 34.2 \\
    & 5.0 & 37.2 & 20.2 & 24.7 & 10.5 \\
    \bottomrule
  \end{tabular}
\end{table}

\textbf{Impact of the Visual Gating Margin ($m$):} 
The margin $m$ directly controls the strictness of the candidate-specific dynamic threshold $\tau_i$, dictating how much local visual similarity is allowed to boost the text-driven logits. As observed in Table~\ref{tab:parameter_sensitivity}, removing the margin entirely ($m=0$) leads to a severe performance collapse (e.g., Surgery CK drops to 23.4). This occurs because a completely loose threshold fails to filter out background noise or uninformative textures, leading to massive false-positive visual boosts that exacerbate semantic ambiguity. Conversely, setting the margin too high ($m=0.4$) overly restricts the visual prior, preventing valid visual clues from calibrating boundary objects. The optimal balance is achieved when $m \in [0.2, 0.3]$, where the gate effectively isolates high-confidence visual similarities.

\textbf{Impact of the Modulation Coefficient ($\alpha$):} 
The coefficient $\alpha$ balances the contribution of the injected visual prior against the original text-driven logits. To ensure that the visual boost is consistent with the logit distribution of the pre-trained foundation model, we introduce an absolute scaling factor $S_{scale}$, adopting the intrinsic temperature $\tau_{OWL-ViT}$ of the foundation model. Because this scaling operation explicitly aligns the magnitude of the visual and textual modalities, the framework is highly robust to variations in $\alpha$, provided it is not set to extreme values. 
However, when $\alpha$ is set excessively large (e.g., $\alpha=5$), performance drastically plummets (Aquatic CK drops to 20.2). This is because an overwhelming reliance on the visual modality overshadows the zero-shot generalization capabilities of the frozen vision-language foundation model, causing the network to overfit to the visual prototypes of known categories and lose its open-vocabulary reasoning ability. Setting $\alpha = 0.5$ provides a robust trade-off, ensuring that the visual prior acts as a helpful calibrator rather than a disruptive override.

\subsubsection{Sensitivity Analysis for UGUOE}
\label{sec:param_uguoe}

\begin{table}[htbp]
  \caption{Parameter sensitivity analysis of the UGUOE module. We
evaluate the local neighborhood size $K_u$, sensitivity factor
$\gamma$, and activation threshold $\tau_{unk}$. Default parameter
values are highlighted in bold in the \emph{Value} column, while the
best metric values within each parameter sweep are highlighted in
bold. U and K denote U-mAP and K-mAP under Task~1, respectively.}
  \label{tab:uguoe_sensitivity}
  \centering
  \small
  \begin{tabular}{cc|cc|cc}
    \toprule
    \multirow{2}{*}{\textbf{Parameter}} & \multirow{2}{*}{\textbf{Value}} & \multicolumn{2}{c|}{\textbf{Aquatic}} & \multicolumn{2}{c}{\textbf{Surgery}} \\
    & & \cellcolor{blue!10}U & \cellcolor{orange!10}K & \cellcolor{blue!10}U & \cellcolor{orange!10}K \\
    \midrule
    \multirow{4}{*}{\shortstack{Local Size\\($K_u$)}} 
    & 1 & \textbf{20.4} & \textbf{63.0} & 16.4 & 43.4 \\
    & \textbf{3} & 20.1 & \textbf{63.0} & \textbf{25.7} & 43.9 \\
    & All & 17.4 & \textbf{63.0} & 18.7 & \textbf{44.0} \\
    & Half & 18.8 & 62.9 & 21.5 & \textbf{44.0} \\
    \midrule
    \multirow{3}{*}{\shortstack{Sensitivity\\($\gamma$)}} 
    & 10 & \textbf{20.2} & 62.8 & 24.0 & \textbf{43.9} \\
    & \textbf{100} & 20.1 & \textbf{63.0} & 25.7 & \textbf{43.9} \\
    & 200 & 18.2 & \textbf{63.0} & \textbf{28.1} & \textbf{43.9} \\
    \midrule
    \multirow{2}{*}{\shortstack{Threshold\\($\tau_{unk}$)}} 
    & \textbf{0.25} & \textbf{20.1} & \textbf{63.0} & 25.7 & \textbf{43.9} \\
    & 0.5 & 18.3 & 62.9 & \textbf{27.5} & 43.5 \\
    \bottomrule
  \end{tabular}
\end{table}

\textbf{Impact of Local Visual Response Aggregation ($K_u$):} The parameter $K_u$ defines the extent of the local semantic neighborhood used to calculate $\mu_{unk}$ and $\sigma^2(S_{local})$. As shown in Table~\ref{tab:uguoe_sensitivity}, using a minimal neighborhood ($K_u=1$) achieves the highest U-mAP on the Aquatic dataset but performs poorly on the complex Surgery dataset due to insufficient uncertainty estimation. Notably, we provide an adaptive alternative referred to as \textbf{Half}, where $K_u$ is set to half the number of known classes. This strategy yields a competitive performance balance (e.g., 21.5 U-mAP on Surgery), offering a robust option that does not rely on a fixed constant across different class pool sizes.

\textbf{Sensitivity to Uncertainty Factor ($\gamma$):} The factor $\gamma$ modulates the penalty for high-certainty regions. We observe that as $\gamma$ increases to 200, the detection performance for unknown objects in the Surgery dataset is further enhanced, reaching its peak at \textbf{28.1} U-mAP. This suggests that more aggressive penalization of "certain" known categories helps the model better identify ambiguous targets in fine-grained medical scenarios. Despite this localized improvement, the overall performance remains relatively stable across a wide range of values ($10 \leq \gamma \leq 200$), demonstrating the robustness of our uncertainty-guided weighting mechanism.

\textbf{Impact of Activation Threshold ($\tau_{unk}$):} The threshold $\tau_{unk}$ acts as a filter for background distractors. While a higher threshold ($\tau_{unk}=0.5$) slightly improves the discovery of unknown tools in the Surgery dataset by suppressing low-confidence artifacts, the default value of 0.25 remains more effective for the Aquatic domain. This trade-off confirms that a moderate threshold is generally sufficient to prevent universal objectness from over-activating in semantic vacuums.

\subsection{Advanced Visualizations and In-depth Analysis}
\label{sec:visualizations}

\subsubsection{Evolution of Known and Unknown Logit Distributions}
\label{sec:score_distribution}

\begin{figure*}[htbp]
  \centering
  \includegraphics[width=0.7\textwidth]{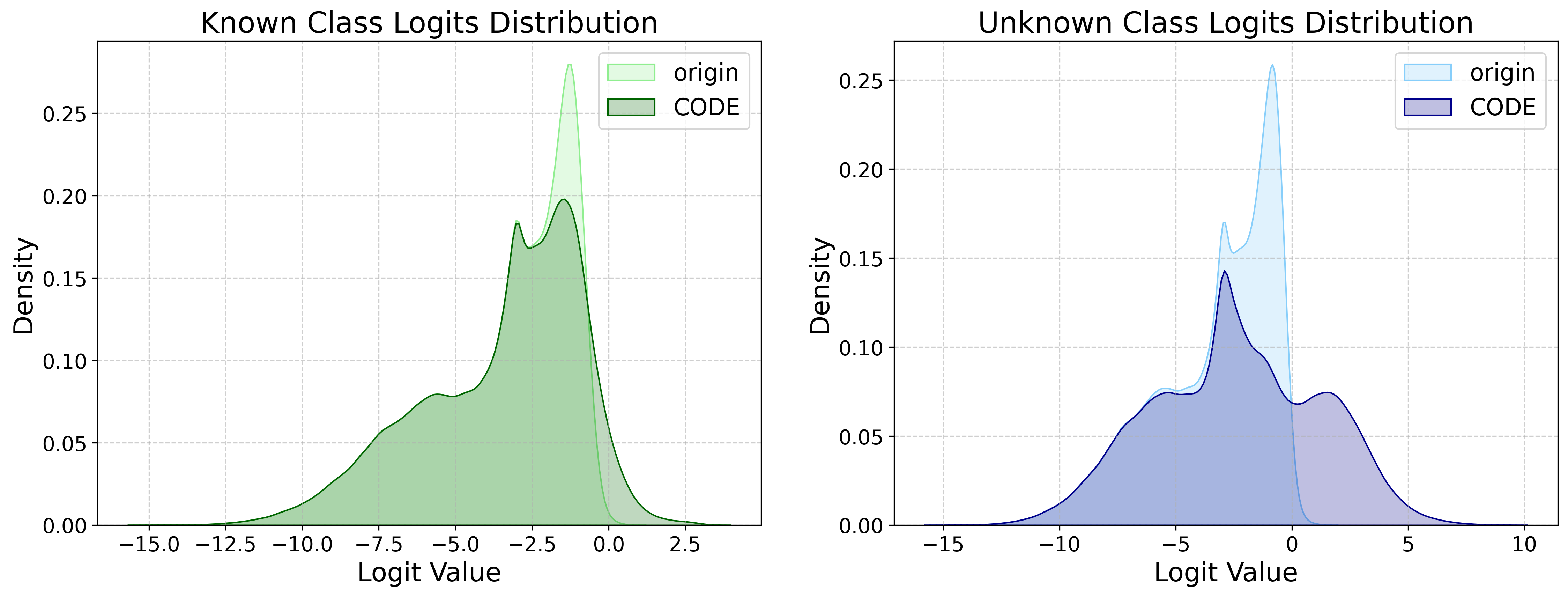}
  \vspace{0.2cm}
  
  \includegraphics[width=0.7\textwidth]{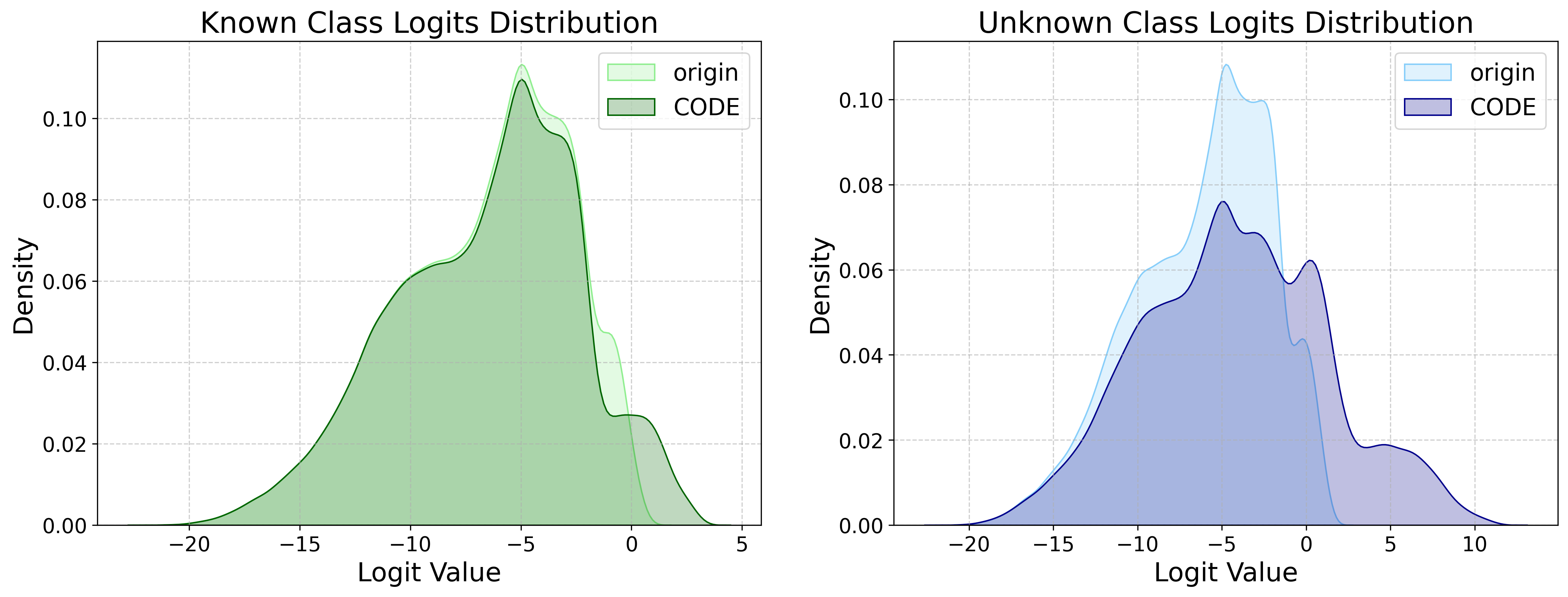}
  \vspace{0.2cm}
  
  \includegraphics[width=0.7\textwidth]{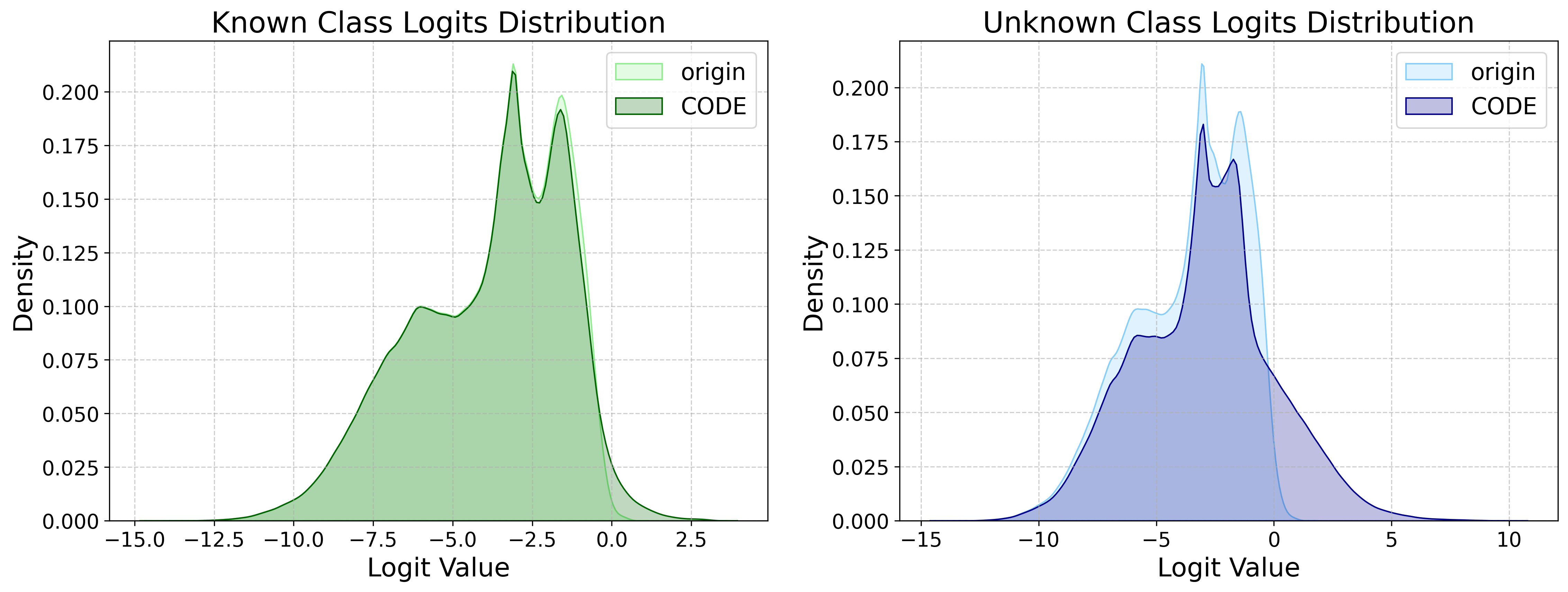}
  \caption{Logit distribution comparisons on the Aquatic (top), Medical (middle), and Surgery (bottom) datasets. For each dataset, the left plot illustrates the known logit distribution, while the right plot displays the unknown logit distribution. CODE significantly shifts the probability mass rightward into the high-confidence region ($>0$).}
  \label{fig:score_dist}
\end{figure*}

To intuitively understand how our proposed framework refines the semantic space and rescues targets from being erroneously filtered, we visualize the logit distributions of both known and unknown targets before and after applying our method. Figure~\ref{fig:score_dist} illustrates these distribution changes across three challenging scenarios: the Aquatic, Medical, and Surgery datasets.

\textbf{Distribution Analysis and Logit Enhancement:} 
Under the previous purely attribute-driven paradigms, a large number of targets suffer from relatively low attribute similarity due to semantic gaps, occlusions, or complex backgrounds. Consequently, their logit distributions are predominantly trapped in the negative region ($< 0$), indicating insufficient confidence for reliable detection. 

In contrast, our CODE framework explicitly enhances reliable target confidence through the synergistic application of the CMJCC and UGUOE modules. As depicted in Figure~\ref{fig:score_dist}, our method drives a conspicuous rightward distribution shift for both known and unknown objects. By paying special attention to the difference density in the positive region (logits $> 0$), it is evident that a massive amount of probability mass has been successfully pushed across the activation threshold. This fundamental logit enhancement directly contributes to the simultaneous improvements in both Known AP and Unknown AP.

\textbf{Insight into the Surgery Dataset:}
Furthermore, these visualizations offer an intuitive explanation for a specific empirical observation in the Surgery dataset. As shown in the bottom row of Figure~\ref{fig:score_dist}, while the unknown distribution in Surgery is drastically enhanced, its known distribution exhibits extremely minimal changes before and after the application of our method—especially when compared to the pronounced known-shifts seen in the Aquatic and Medical datasets. This marginal enhancement in the known distribution, coupled with the
aggressive rightward expansion of the unknown distribution, is
consistent with the relatively weaker known-class performance on
Surgery; in the final main comparison, CODE is $1.7$ K-mAP points
below PASS on this dataset. It suggests that the massive activation of ambiguous unknown surgical tools introduces a minor but unavoidable interference into the previously established known decision boundaries.

\subsubsection{Visualization of Dynamic Outlier Suppression (DOSCM)}
\label{sec:doscm_vis}

\begin{figure*}[htbp]
  \centering
  \includegraphics[width=0.3\textwidth]{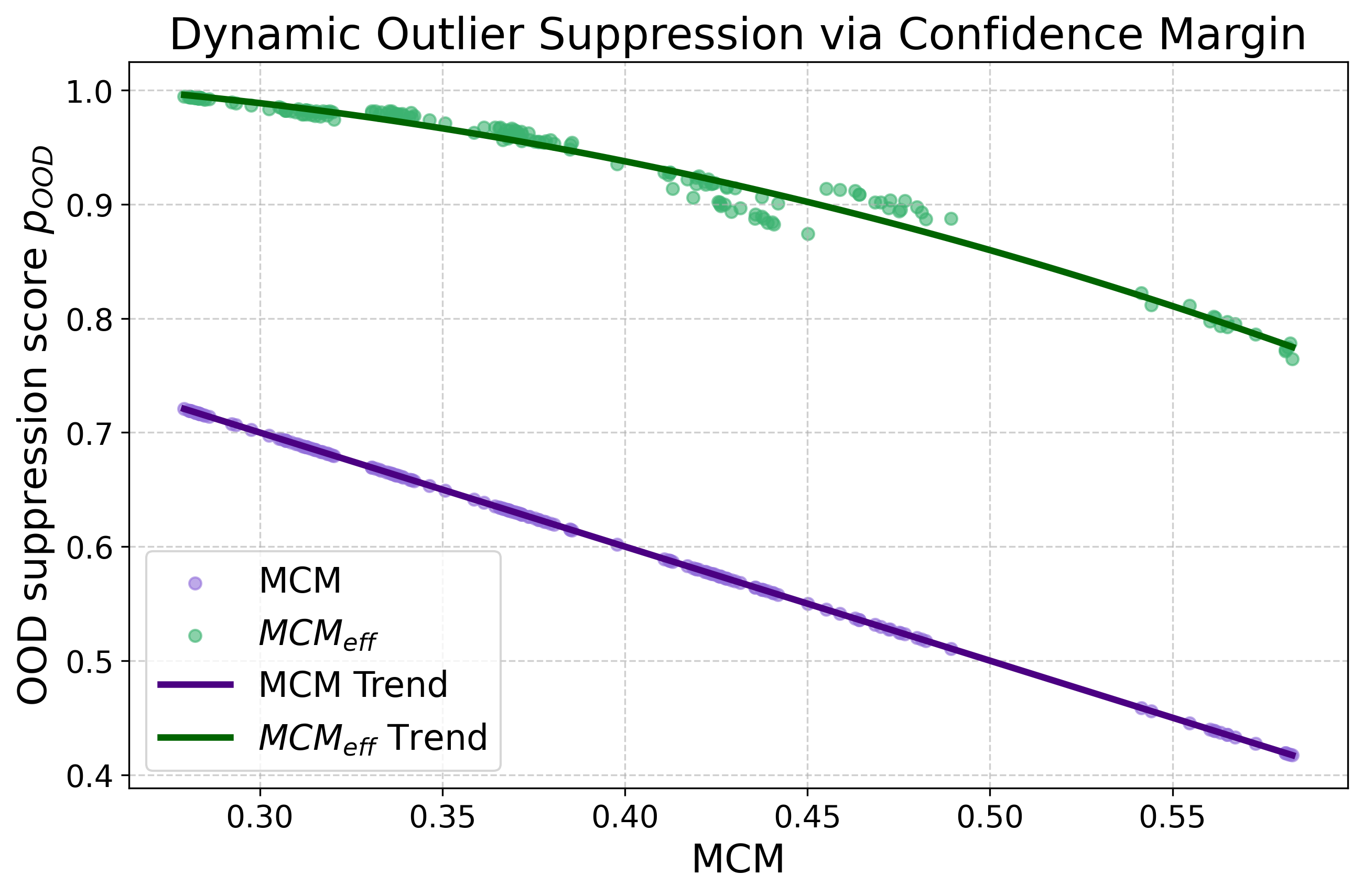}
  \hfill
  \includegraphics[width=0.3\textwidth]{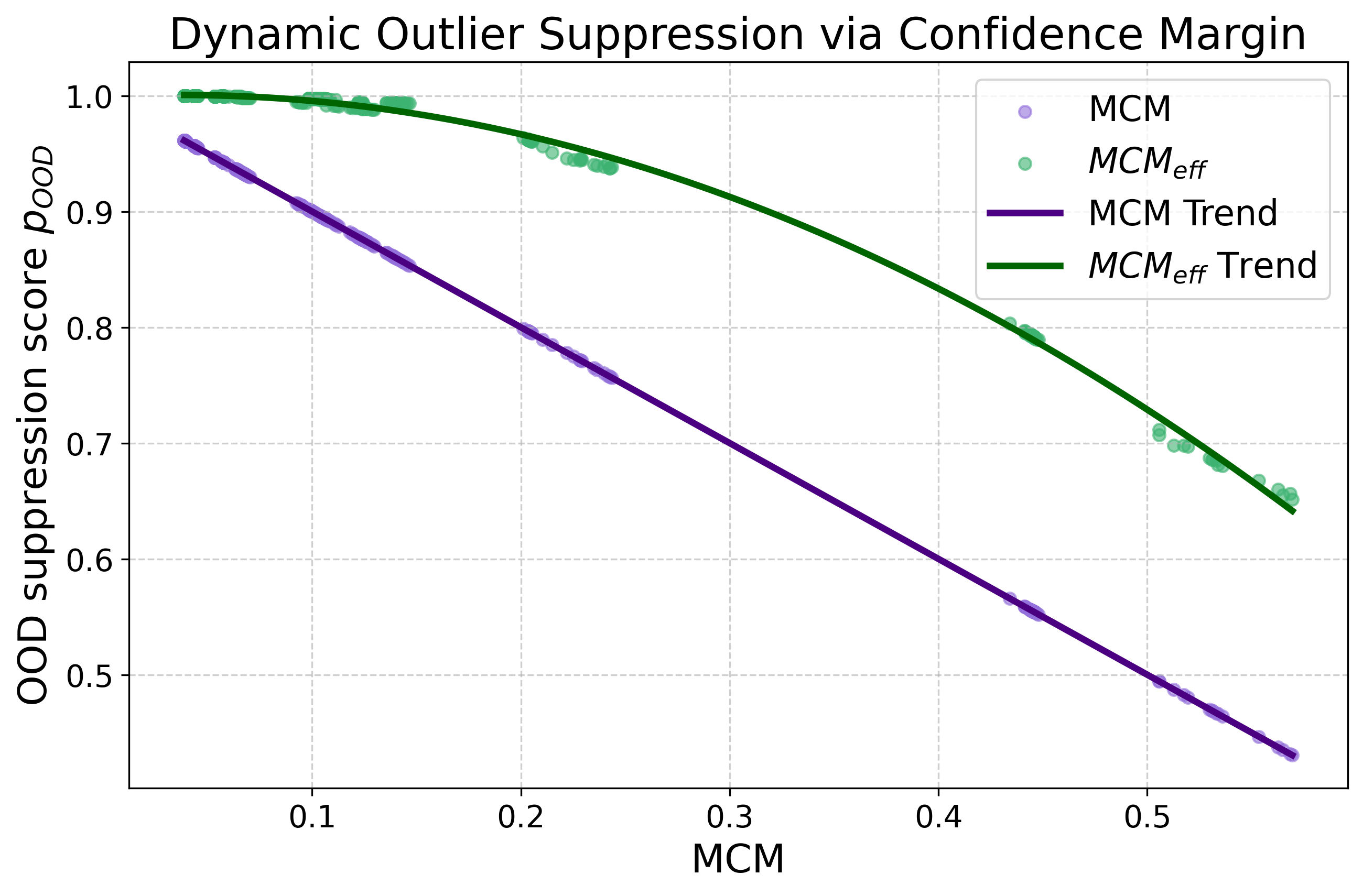}
  \hfill
  \includegraphics[width=0.3\textwidth]{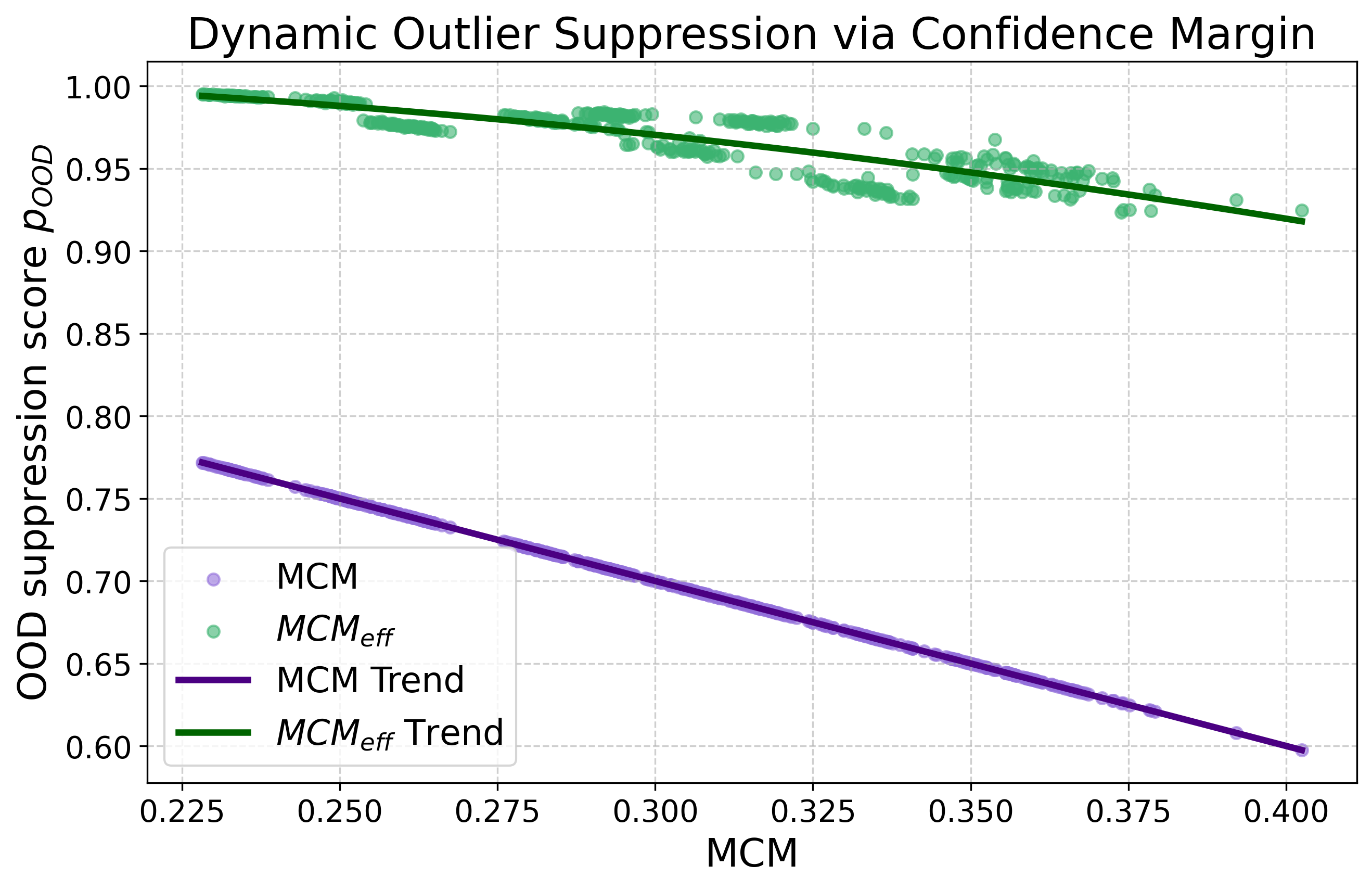}
  \caption{Scatter visualizations of DOSCM on the Aquatic (left), Game (middle), and Surgery (right) datasets. The x-axis represents the Top-1 known Softmax probability, and the y-axis is the final OOD score ($p_{OOD}$). The purple line denotes the rigid baseline suppression bottleneck, while the green curve illustrates our dynamic suppression boundary.}
  \label{fig:doscm_scatter}
\end{figure*}

To intuitively demonstrate how the DOSCM module protects boundary unknown targets and shapes the decision boundary, we visualize the correlation between the semantic confidence margin ($\Delta S$) and the effective suppression penalty in Figure~\ref{fig:doscm_scatter}.

\textbf{Analysis:} As illustrated in the scatter plots, the x-axis represents the Top-1 known Softmax probability (MCM base), and the y-axis is the final Out-of-Distribution score ($p_{OOD}$). The \textcolor{purple}{\textbf{purple line}} represents the rigid, linear suppression bottleneck of the baseline MCM, which forcefully penalizes boundary unknown objects, misclassifying them as background simply because they share visual traits with a known class. 

In contrast, the \textcolor{green}{\textbf{green curve}} represents the dynamic suppression boundary generated by DOSCM. By modulating the penalty based on the exclusivity gap $\Delta S$, DOSCM dynamically shrinks the penalty for ambiguous targets. Consequently, their scatter points are successfully pulled upward from the rigid purple baseline, preserving their enhanced unknown probabilities. This dynamic rescue mechanism is remarkably vital in complex domains like Surgery (Figure~\ref{fig:doscm_scatter}, right), where it safely lifts a dense cluster of hard medical instruments above the suppression threshold.

\textbf{Insight into the Game Dataset (False Positives):}
However, this dynamic relaxation introduces an inherent trade-off in certain highly synthetic domains. As observed in the Game dataset (Figure~\ref{fig:doscm_scatter}, middle), which consists of synthetic avatars with extreme inter-class similarity, lifting the suppression boundary rescues hard unknowns but simultaneously elevates the OOD scores of some ambiguous known objects. This over-relaxation allows these known objects to cross the unknown decision boundary, resulting in an increase in unknown false positives (i.e., genuine known objects being misclassified as unknowns). This observation is also consistent with the feature-space analysis
reported in the main paper, where Game exhibits an average inter-class
cosine similarity of $0.82$. This visual phenomenon provides a clear mathematical explanation for why the CODE framework experiences a performance degradation in Unknown AP (U-mAP) on the Game dataset, highlighting the unique challenges of dynamic margin tuning in synthetic environments.

\subsubsection{Qualitative Analysis on Task 1 Discovery}
\label{sec:qualitative_task1}

\begin{figure*}[htbp]
  \centering
  \includegraphics[width=\textwidth]{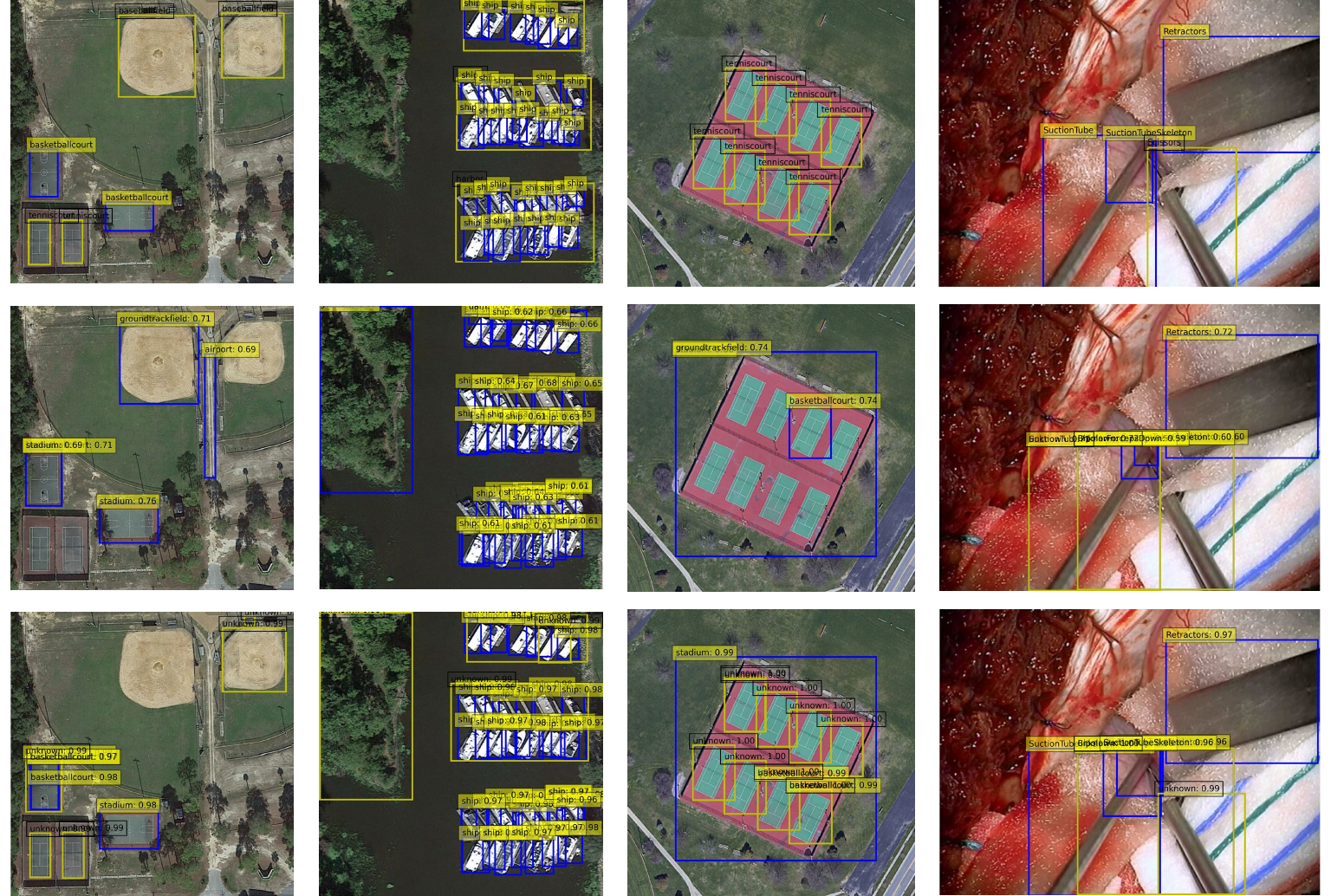}
  \caption{Qualitative detection results on the Aerial and Surgery datasets for Task 1. Top row: Ground Truth annotations. Middle row: Predictions by the SOTA baseline PASS. Bottom row: Predictions by our proposed CODE framework. Blue boxes denote known categories, while yellow boxes denote unknown categories.}
  \label{fig:qualitative_task1}
\end{figure*}

To further evaluate the discovery capability of our framework, we provide a qualitative comparison between the baseline PASS \cite{yang2025detecting} and our proposed CODE on the Aerial and Surgery datasets under Task 1. As shown in Figure~\ref{fig:qualitative_task1}, CODE demonstrates a significant advantage in recalling challenging unknown objects that are frequently missed by existing SOTA methods.

\textbf{Analysis of Recalled Unknown Objects:}
By comparing the bottom row (CODE) with the middle row (PASS) from left to right, we observe that CODE successfully recovers several critical unknown instances where the baseline fails. For instance, in the first and third columns of the Aerial scenario, PASS misses the ``Tennis Court'' because it shares high semantic and geometric similarity with the known ``Basketball Court'' category, causing it to be either misclassified or suppressed. Our UGUOE module effectively identifies the model's classification hesitation in this semantic vacuum, while DOSCM provides a dynamic margin to protect these boundary targets. Similarly, in the middle column, CODE successfully recalls the ``Harbor''—a target typically cluttered with numerous ``Ship'' instances that cause severe occlusion and visual complexity. By leveraging visual prototypes and uncertainty-guided enhancement, our method distinguishes the harbor's global context from the dense local ship responses. Furthermore, in the surgical scenario (right column), CODE successfully discovers the unknown ``Scissors''. This object is particularly challenging as its metallic texture and slender structure closely resemble the known ``Suction Tube'' category. While the rigid suppression mechanism in the baseline tends to filter such ambiguous instances, CODE maintains a high objectness response for the scissors by quantifying the local visual variance and applying a soft confidence margin. Ultimately, these qualitative results provide strong visual evidence that the synergy of UGUOE and DOSCM effectively mitigates the rigid over-suppression bottleneck, ensuring a much higher recall for hard-to-detect objects in complex real-world environments.

\begin{figure*}[htbp]
  \centering
  \includegraphics[width=\textwidth]{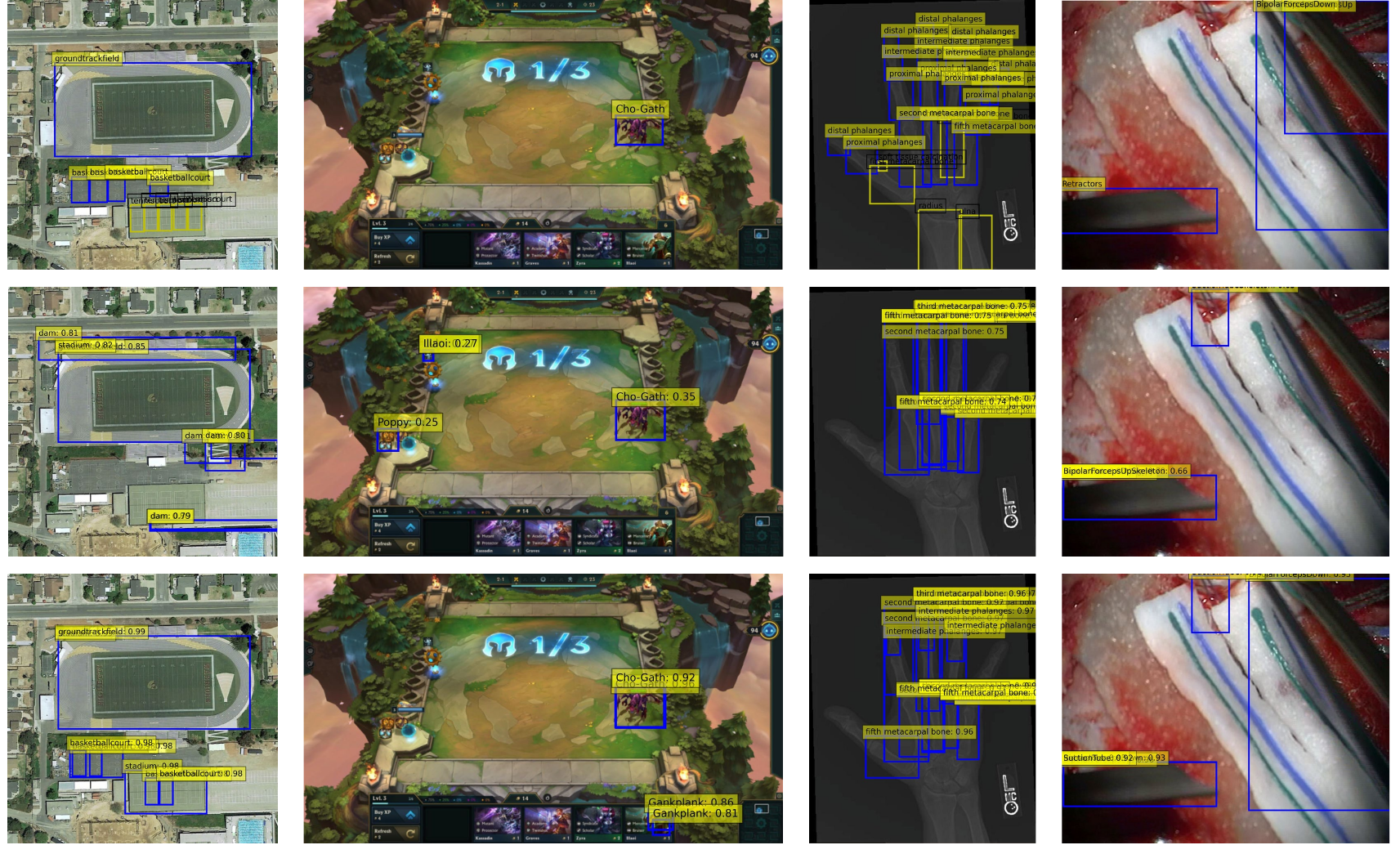}
  \caption{Qualitative detection results on Task 2 across the Aerial, Game, Medical, and Surgery datasets. Top row: Ground Truth annotations, where yellow boxes indicate Currently Known (CK) classes. Middle row: Predictions by the baseline PASS. Bottom row: Predictions by our CODE framework. CODE recalls significantly more known targets (blue boxes) through visual-to-visual calibration.}
  \label{fig:qualitative_vis3}
\end{figure*}

\subsubsection{Qualitative Analysis on Task 2}
\label{sec:qualitative_task2}

In the Task~2 evaluation, we analyze the model's detection performance
after joint retraining on the expanded category set containing the
categories from both Task~1 and Task~2. As shown in Figure~\ref{fig:qualitative_vis3}, the ground-truth annotations for the CK (Currently Known) categories are denoted by yellow boxes, and the model is expected to predict them as known categories (blue boxes). The results demonstrate that, benefiting from the visual reference provided by the Cross-Modal Joint Confidence Calibration (CMJCC) module, the CODE framework significantly improves the recall rate of these targets.

Specifically, in the first column (Aerial), CODE successfully identifies more ``Basketball Court'' instances than the baseline. However, due to the extremely high geometric and semantic overlap between tennis courts and basketball courts, some omissions of the ``Tennis Court'' still occur. In the second column (Game), our framework correctly identifies the known class ``Cho-Gath''; however, as previously analyzed, it hallucinates an incorrect known category nearby. This reflects the challenge of precise semantic differentiation in synthetic environments characterized by extreme inter-class visual similarity. In the third column (Medical), CODE effectively localizes more targets than the baseline, including various ``Metacarpal Bones'' and ``Phalanges,'' although the ``Radius'' is still missed. Finally, in the fourth column (Surgery), CODE accurately identifies the ``Bipolar Forceps Down'' category, which is missed by the baseline PASS. The baseline's failure is primarily due to the inconspicuous attribute features of these tools, which makes the pure text-driven logic difficult to activate effectively. CMJCC successfully compensates for this lack of visual-textual alignment by injecting global visual prototypes, ensuring the successful recall of the targets.

In summary, these visualization results confirm that by aggregating global visual prototypes and injecting them into the inference pipeline, CMJCC provides a robust visual calibration reference. Although challenges remain when dealing with objects exhibiting extreme inter-class similarity or significant scale variations, the improvement in known class recall across multiple domains in Task 2 proves the effectiveness of our cross-modal joint calibration strategy when handling the full set of known data.

\subsubsection{Qualitative Analysis on Class Confusion and Limitations}
\label{sec:qualitative_confusion_errors}

\begin{figure*}[htbp]
  \centering
  \includegraphics[width=\textwidth]{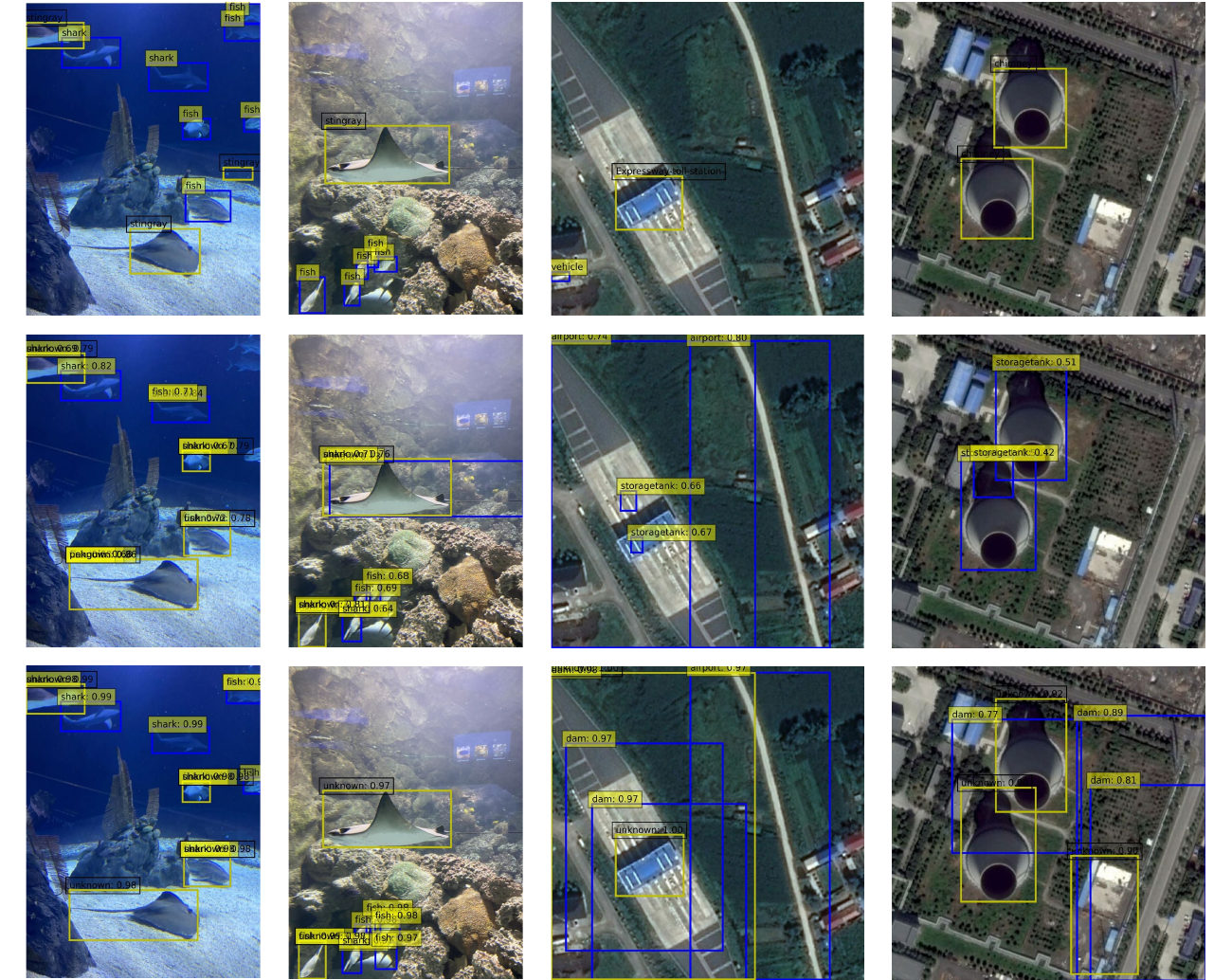}
  \caption{Qualitative detection results focusing on class confusion and error cases on the Aquatic and Aerial datasets for Task 1. Top row: Ground Truth annotations. Middle row: Predictions by the SOTA baseline PASS. Bottom row: Predictions by our proposed CODE framework. Blue boxes denote known categories, while yellow boxes denote unknown categories.}
  \label{fig:qualitative_vis2}
\end{figure*}

To provide a comprehensive and transparent evaluation, we further analyze scenarios involving severe semantic confusion and discuss the inherent limitations of our framework. Figure~\ref{fig:qualitative_vis2} presents a qualitative comparison on the Aquatic (left two columns) and Aerial (right two columns) datasets under Task 1.

\textbf{Mitigating Semantic Confusion:}
A major challenge in OWOD is the semantic overlap between novel targets and established categories. As observed in the second and fourth columns of Figure~\ref{fig:qualitative_vis2}, the baseline method frequently struggles with this confusion. For the unknown ``Stingray'' (second column), the baseline predicts both an unknown bounding box and a hallucinated known box, indicating a failure to decisively separate the semantic spaces. In contrast, CODE elegantly resolves this confusion, isolating the target purely as unknown without overlapping known predictions. Similarly, in the fourth column, the baseline rigidly misclassifies the unknown ``Chimney'' as a known ``Storage Tank'' due to their cylindrical geometric similarities. By leveraging UGUOE to capture classification hesitation and applying DOSCM to adjust the suppression boundary, CODE successfully rescues the chimney into the correct unknown category.

\textbf{Error Analysis and Limitations:}
Despite these robust advantages, the dynamic nature of our modules introduces specific trade-offs, leading to certain failure cases shown in the first and third columns. 
\begin{itemize}
    \item \textbf{False Positives on Unknowns (Aquatic):} In the first column, while CODE successfully recalls more targets that blend seamlessly into the complex background (e.g., the fish in the top right corner missed by the baseline), it incorrectly classifies a genuine known fish as an unknown object. This over-activation is a side effect of the UGUOE module. When a known object shares intense visual similarities with the background, its local visual variance increases, causing the uncertainty-guided mechanism to misinterpret it as an ambiguous target and erroneously assign it an unknown label.
    \item \textbf{Hallucination of Known Classes (Aerial):} In the third column, CODE correctly discovers the complex unknown ``Expressway Toll Station''. However, it simultaneously produces a false positive prediction for the known ``Dam'' category on the same structure. We attribute this to the Cross-Modal Joint Confidence Calibration (CMJCC) module. While the injection of visual prototypes effectively calibrates and enhances the recall of known classes, it may inadvertently over-boost the known logits of background regions or unknown targets that share intense textural or structural similarities with the known prototypes. 
\end{itemize}

These failure cases highlight the delicate balance required in Open World Object Detection. While dynamic margins and uncertainty-guided enhancements significantly boost overall performance and unknown recall, refining the cross-modal calibration to prevent over-boosting in extreme boundary cases remains an important direction for future research.

\section{Limitations and Future Work}
\label{sec:limitations_future_work}

\textbf{Limitations:} 
Despite the improvements achieved by the CODE framework, several limitations remain. As discussed in Sec.~4.5 of the main paper, the few-shot experiment
isolates the effect of sample scarcity but does not constitute a
complete evaluation under long-tailed class distributions or noisy
annotations. Moreover, because the visual prototypes are
dataset-specific estimates, substantial cross-domain distribution
shift may reduce their representativeness. First, semantic aliasing persists in extreme domains; as observed in the Game and Surgery datasets, high inter-class visual similarity can cause the UGUOE module to over-boost ambiguous regions, leading to unknown false positives. Second, the granularity of visual prototypes is currently limited, as representing each category with a single centroid may fail to capture intra-class diversity such as variations in viewpoint or occlusion. Finally, the framework is subject to a fundamental dependency on pre-trained models. Since CODE functions as a calibration layer atop frozen vision-language foundation models like OWL-ViT, its overall performance ceiling is inherently constrained by the zero-shot generalization and inherent discriminative capabilities of these underlying models.

\textbf{Future Work:} 
For future research, we identify several promising directions. We aim to explore dynamic and multi-prototype representations by incorporating clustering techniques (e.g., K-Means) to maintain multiple prototypes per category, thereby enhancing the model's robustness to intra-class variance. While the current CODE framework performs attribute generation
offline, we further consider Dynamic Attribute Generation as a future
direction. Instead of relying on static, offline attribute pools, this approach would leverage Large Multimodal Models (LMMs) to generate descriptive and judgmental attributes in real-time, conditioned on the specific visual context of the input image. Finally, we intend to investigate cross-domain adaptation for OWOD. By integrating unsupervised domain adaptation (UDA) techniques, the framework could better generalize across the diverse data distributions of the RWD benchmark without requiring exhaustive few-shot exemplars for every novel scenario, ultimately moving towards more autonomous and life-long open-world perception.

\end{document}